\documentclass[lettersize,journal]{IEEEtran}
\usepackage{amsmath,amsfonts}
\usepackage{algorithm, algorithmic}
\usepackage{array}
\usepackage[caption=false,font=normalsize,labelfont=sf,textfont=sf]{subfig}
\usepackage{textcomp}
\usepackage{stfloats}
\usepackage{url}
\usepackage{verbatim}
\usepackage{graphicx}
\def\BibTeX{{\rm B\kern-.05em{\sc i\kern-.025em b}\kern-.08em
    T\kern-.1667em\lower.7ex\hbox{E}\kern-.125emX}}
\usepackage{balance}
\begin{document}
\title{Unsupervised Ensemble Methods for Anomaly Detection in PLC-based Process Control}
\author{Emmanuel Aboah Boateng,  \textit{Student Member, IEEE}, and J.W. Bruce, \textit{Senior Member, IEEE}
\thanks{

The authors are with the Department of Electrical and Computer Engineering, Tennessee Technological University, Cookeville, TN 38505 USA (email: jwbruce@tntech.edu)

The opinions expressed here are entirely that of the author. No warranty is expressed or implied. User assumes all risk.
}
}


\maketitle

\begin{abstract}
Programmable logic controller (PLC) based industrial control systems (ICS) are used to monitor and control critical infrastructure. Integration of communication networks and an Internet of Things approach in ICS has increased ICS vulnerability to cyber-attacks. This work proposes novel unsupervised machine learning ensemble methods for anomaly detection in PLC-based ICS. The work presents two broad approaches to anomaly detection: a weighted voting ensemble approach with a learning algorithm based on coefficient of determination and a stacking-based ensemble approach using isolation forest meta-detector. The two ensemble methods were analyzed via an open-source PLC-based ICS subjected to multiple attack scenarios as a case study. The work considers four different learning models for the weighted voting ensemble method. Comparative performance analyses of five ensemble methods driven diverse base detectors are presented. Results show that stacking-based ensemble method using isolation forest meta-detector achieves superior performance to previous work on all performance metrics. Results also suggest that effective unsupervised ensemble methods, such as stacking-based ensemble having isolation forest meta-detector, can robustly detect anomalies in arbitrary ICS datasets. Finally, the presented results were validated by using statistical hypothesis tests.
\end{abstract}


\section{Introduction}
\IEEEPARstart{I}{ntegrating} physical control processes, computing, and communication through Internet of Things technology (IoT) improves industrial control systems (ICSs) production efficiency, flexibility, and reliability. Unfortunately, the prevalence of Internet of Things technology and networked sensors in many ICSs opens up opportunities for cybercriminals to leverage ICS vulnerabilities towards initiating cyber-attacks \cite{sakhnini2021security, haddadpajouh2021survey}. There has been an increase in cyber-attack awareness of critical infrastructure \cite{kello2019virtual, yaacoub2020cyber, thakur2016impact, pleta2020cyber}. Programmable Logic Controllers (PLCs) are industrial computers that play a significant role in ICS. PLCs are a family of embedded devices critical to ICS network operation. PLCs collect input data from field devices such as sensors and send output commands to actuating devices to control ICS operations \cite{wardak2016plc, abbasi2017ecfi}. ICS plays a vital role in monitoring and controlling critical infrastructures such as electricity supply, nuclear plants, petrochemical industries, and water management. PLCs, which serve as the heart of ICS, are vulnerable to attacks like other embedded devices. Because PLCs are broadly used to control the physical processes of critical infrastructure in societies, PLC malfunctions or attacks can cause physical and economic damages as well as compromise the safety of ICS service personnel \cite{abbasi2016ghost}.

Traditionally, PLCs use proprietary hardware and software in physically secure locations with no external connection to the internet \cite{yau2015plc, langmann2019plc}. As a result, attacks on PLCs were limited to insider intrusion, physical damage, and tampering with PLC devices \cite{tsiknas2021cyber}. In recent times, PLCs have adopted corporate networks and common information technologies with integrated smart sensors and wireless networks for performance enhancement, and efficiency improvement \cite{spyridopoulos2013incident, hwang2021sfd}. Applying traditional techniques and contemporary tools for detecting PLC anomalous behavior is difficult due to their unique architecture and proprietary operating systems. Therefore, it is crucial to protect PLCs against any forms of attack or anomalies such as hardware malfunction, accidental actions by insiders, and malicious intruders \cite{boeckl2019considerations}.

Machine learning techniques have been applied for anomaly detection in embedded devices and for identifying anomalies in several computing applications \cite{chen2017application, welborn2021one, hiranai2021detection, shitharth2017enhanced, yin2017deep, aboah_access}. Recently, work has been done that utilizes ICS data to develop unsupervised anomaly detection models which exhibit the ICS system's normal behavior and identify all different behaviors as anomalies \cite{aboah2022plc, elnour2020hybrid, boateng2021anomaly}. These algorithms learn a single hypothesis from training dataset in order to make generalizations about unseen events. Moreover, different anomaly detection algorithms make different assumptions about the dataset, which may not be accurate. As a result, different anomaly detection algorithms perform well on some subsets of a dataset and perform poorly on other subsets \cite{aggarwal2017introduction}. Anomaly detection algorithms' unique assumptions about datasets make it challenging to generalize their performance to an arbitrary dataset. 

Ensemble methods refer to the approach by which multiple models are strategically developed and combined for improved performance. In supervised machine learning classification domain, ensemble methods broadly include bagging, boosting, stacking \cite{aggarwal2014algorithms, aggarwal2015data}. However, in unsupervised outlier analysis domain, there are two primary types of ensembles, namely sequential ensembles and independent ensembles \cite{aggarwal2017introduction}. In sequential ensembles, a given algorithm or set of algorithms are applied sequentially such that previous applications influence future applications of the algorithm. In contrast, different anomaly detection algorithms are applied to either the complete dataset or portions of the dataset in independent ensembles. In \cite{aggarwal2017introduction}, Aggarwal claims that ensemble combinations of multiple anomaly detection algorithms are often able to perform more robustly on arbitrary datasets due to their ability to combine the strengths of multiple detection algorithms. The work in \cite{aboah2022plc} suggests that ensemble learning techniques may improve anomaly detection in ICS based on the performances of several anomaly detection algorithms. Several studies have focused on improving supervised classification performance using homogeneous classifiers \cite{cabrera2008ensemble, folino2016distributed}, heterogeneous classifiers \cite{zhao2015ensemble}, or a combination of both \cite{amozegar2016ensemble, aburomman2016novel}. However, there is not sufficient study concerning unsupervised ensemble machine learning techniques for anomaly detection in PLCs and ICSs. 

This work proposes five generalized unsupervised ensemble techniques for ICS anomaly detection: majority voting, maximum-score, soft voting, weighted voting, and stacking-based anomaly detection approaches. The weighted voting ensemble method introduces a new approach for assigning weights to base detectors based on their coefficient of determination. The stacking-based approach is a new ensemble approach based on IF meta-learner. The proposed methods broadly fall under independent ensembles in outlier analysis. For each ensemble method, this work adopts the same base detectors, namely; OCSVM, OCNN, and IF, that learn new representations from the training dataset. Then, output representations from the base detectors are concatenated and passed to a voting strategy algorithm to detect attacks from the newly merged representations. The trained models are intended to run on a dedicated or separate computer to monitor operations at the PLC memory addresses through real-time HMI historian logs. Experimental results show that the proposed ensemble techniques in this work outperform existing approaches with acceptable statistical significance.

In summary, the main contributions of this work are listed as follows:
\begin{enumerate}
\item	Propose first known ensemble anomaly detection methods using OCSVM, OCNN, and IF base detectors;
\item   Introduce a weighted voting ensemble technique with a learning algorithm for assigning weights to base detectors based on their coefficient of determination; and
\item	Introduce a stacking-based anomaly detection framework with isolation forest meta-detector for robust performance.
\end{enumerate}

The remainder of the paper is organized as follows. Section 2 reviews related works. Section 3 presents details of the experiment setup and data collection approach. Section 4 presents the proposed methods, followed by section 5, which presents results and analysis. Section 6 presents the conclusions and recommendations for future work.

\section{Related Work}

\subsection{Unsupervised Ensemble Methods}
Sun et al. \cite{sun2016detecting} employed an unsupervised ensemble outliers detection framework based on IF algorithms to detect insider threats in an industry. They modified the isolation trees in the IF to support categorical data. Although they achieved significant detection recall values, a drawback of the approach in \cite{sun2016detecting} is that the ensemble method consisted only of isolation trees. As a result, the isolation trees made the same assumptions about the dataset.

Haider et al. \cite{haidar2018adaptive} proposed an ensemble method based on OCSVM and IF on data clusters. They employed human domain experts to identify false-positive results, which were used to update the ensemble algorithm progressively. They executed a k-means algorithm for class decomposition on top of the base detectors. Finally, they tuned the ensemble algorithm by oversampling the false-positive results to create an accurate decision boundary. Although this algorithm presents an interesting idea for unsupervised anomaly detection, it requires human intervention to achieve optimal performance. 

In \cite{parveen2013evolving}, the authors proposed an ensemble learning framework that combines an unsupervised learning method with a graph-based anomaly detection technique. The base detectors consisted of OCSVM and minimum description length (MDL)\cite{diop2019design}. They also proposed a detection model solely based on multiple OCSVM and a streaming data mining approach to account for concept drift. Reported performance was was, suggesting that choice and number of base detectors in an ensemble algorithm are crucial to obtain the best results. 

Yuan et al. \cite{yuan2018insider} proposed an ensemble of deep learning algorithms consisting of long short-term memory and a convolutional neural network to identify anomalies in multiple scenarios. They achieved an area under the curve of $94\%$. Although their method worked well, their approach requires significant computational resources like graphical processing units during training and testing. 

\subsection{Base Detectors for Ensemble Methods}
Inoue et al. utilized unsupervised ML algorithms for anomaly detection in water treatment systems \cite{inoue2017anomaly}. They compared a deep neural network consisting of feedforward layers with multiple inputs and outputs with a One-class Support Vector Machine (OCSVM). The authors claimed that the deep neural network model produced fewer false positives than the OCSVM, although the OCSVM could detect more anomalies. The authors achieved recall values of less than 0.7 for both deep neural network and OCSVM. 

In \cite{muna2018identification}, the authors employed a fully connected neural network and an autoencoder to detect anomalies in IoT computer network traffic. Their results showed a higher detection rate and lowered false positive rate when compared with eight other modern anomaly detection techniques. Potluri et al. \cite{potluri2017identifying} proposed using artificial neural networks for identifying false data injection attacks in ICS. The classification report obtained in \cite{potluri2017identifying} shows a promising detection accuracy with artificial neural networks.

Ahmed et al. \cite{ahmed2019unsupervised} presented an unsupervised learning IF approach to detect hidden data integrity assault on a smart grid communication network. They reported an average accuracy of 93\%,using simulated experimental data. While the simulated data may not depict ICS accurately, The work in \cite{ahmed2019unsupervised} suggests that IF has the potential for high performance in anomaly detection.

Although both supervised and unsupervised ML techniques have been applied in PLC anomaly detection  \cite{aboah2022plc, boateng2021anomaly, ahmad2020machine}, it is usually difficult to rely on a supervised learning approach as real-world ICS contain numerous sensor data that are tedious to label. An unsupervised ML technique called OCSVM was employed to detect anomalies in PLCs successfully \cite{aboah2022plc}. In \cite{aboah2022plc}, the authors employed OCSVM, OCNN, and IF for PLC anomaly detection. Their experiment simulated a traffic light control system using a PLC. They captured relevant PLC memory addresses into a log file for real-time data recording of system operations. The captured data was normalized and used to train and evaluate their models. Aboah et al. concluded that IF outperforms OCNN and OCSVM on the TLIGHT system dataset. 

While OCSVM, OCNN, and IF have been used as effective unsupervised techniques for anomaly detection, OCSVM performance is unsatisfactory on complex, high-dimensional datasets \cite{chalapathy2018anomaly, bengio2007scaling, fawcett2003proceedings}. Although OCNN has been used for anomaly detection in a complex dataset \cite{chalapathy2018anomaly, aboah2022plc}, the work in \cite{aboah2022plc} shows that OCNN performs at par with OCSVM on several test cases of their TLIGHT dataset because OCNN and OCSVM are formulated from a similar optimization problem. The work in \cite{aboah2022plc} also shows that some anomalous data points are detected by either OCSVM or OCNN, but IF fails to detect them. Therefore, this work hypothesizes that some base anomaly detectors will perform well on a particular subset of a dataset, whereas other base detectors will perform better on other subsets. This work proposes five unsupervised ensemble techniques based on OCSVM, OCNN, and IF to improve anomaly detection performance.

\section{Experiment Setup}
This section provides the details of the experimental setup for simulating the behavior of the traffic light (TLIGHT) system in \cite{aboah2022plc, siemens1996s7} for the purposes of data collection. The details of the experimental setup is provided in \cite{aboah2022plc}.

The experimental setup is based on Siemen's open-source TLIGHT program described in \cite{aboah2022plc, siemens1996s7}. OpenPLC is an open-source PLC simulation platform for home and industrial automation systems development \cite{alves2014openplc}. OpenPLC editor was used to simulate and test the TLIGHT system logic to ensure that the program was error-free and accurately depicted the TLIGHT system description in \cite{aboah2022plc, siemens1996s7}. The ladder logic was converted to a structured text format and uploaded to OpenPLC runtime for execution.
ScadaBR \cite{mazurkiewicz2016open}, an open-source Supervisory Control and Data Acquisition (SCADA) system, was employed as the HMI to monitor and control the PLC runtime. HMI application runs independently of the PLC. The PLC input and output memory addresses were mapped to corresponding Modbus input and output addresses in the HMI. At the end of every HMI cycle time (100 ms), ScadaBR logs available data at the input and output Modbus addresses to a log file. Finally, TLIGHT system operations are exported from the HMI as comma-separated-values files for preprocessing and machine learning model training.

\section{Proposed Method}
\begin{figure}[!t]
\centering
\includegraphics[width=9 cm]{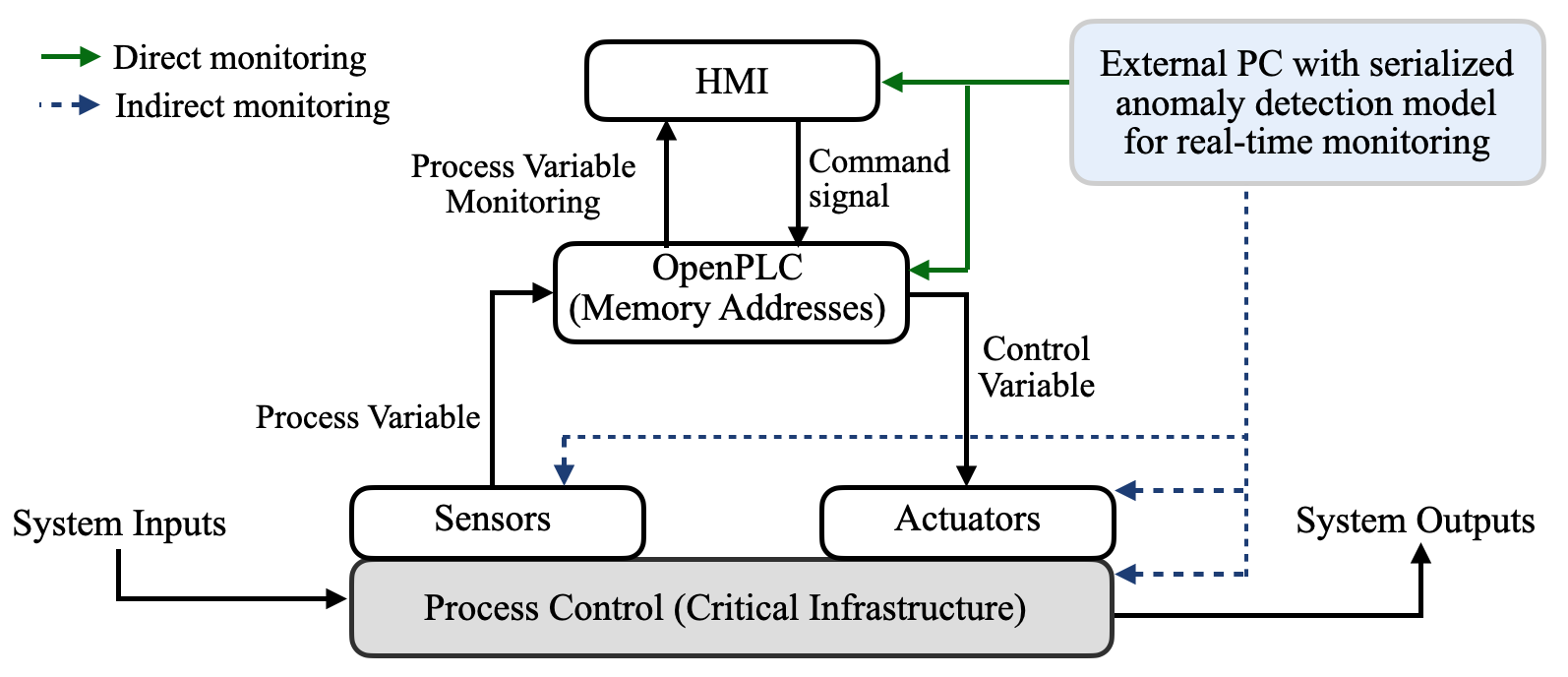}
\caption{Anomaly detection framework}
\label{fig1}
\end{figure}

The proposed unsupervised ensemble methods use the normal process data from the TLIGHT system's input and output signals. This section presents the details about the data collection, anomalies, and theoretical background of the algorithms used in the proposed methods. Figure~\ref{fig1} shows the general anomaly detection framework of the proposed methods. The anomaly detection framework in Figure~\ref{fig1} represents a real-world scenario whereby the trained ensemble models in this work are serialized onto a separate personal computer (PC) for real-time PLC monitoring and anomaly detection. The trained ensemble models receive the process control's real-time data through the HMI and the OpenPLC memory addresses in order to determine whether the received signals violate the process control's normal operations.

In order to evaluate and compare the proposed unsupervised ensemble methods performance in this work, five different test sets are generated. Each test set contains normal and anomalous TLIGHT system events. Anomalous system events for the five test sets are derived from seven scenarios. The attack scenarios considered in the experiment could represent real-world anomalies resulting from malfunctioning sensors and actuators, such as physical obstruction, natural disaster, or malicious attacks by cyber threat actors. The five test cases were considered based on combinations of the anomalous scenarios. The total training dataset samples and test sets 1--5 are identical with the setup used in \cite{aboah2022plc}. Test sets 4 and 5 consist mainly of timing bits anomalies. Table~\ref{tab1} summarizes the number of records and proportion of anomalies in the training and test sets.

\begin{table}
\begin{center}
\caption{Number of records and proportion of anomalies in training and test data sets}
\label{tab1}
\begin{tabular}{| c | c | c |}
\hline
\textbf{Dataset}	& \textbf{No. of Records}	& \textbf{$\textbf{\%}$ Anomalies}\\
\hline
Training set 	& 41580 		& n/a\\
\hline
Test Set 1		& 5000			& 10\\
\hline 
Test Set 2		& 7000			& 10\\
\hline 
Test Set 3		& 13130			& 20\\
\hline 
Test Set 4		& 15000			& 30\\
\hline 
Test Set 5		& 18270			& 50\\
\hline 
\end{tabular}
\end{center}
\end{table}

\subsection{Base Detectors}
Base detectors refer to the individual components of an ensemble technique. The base detectors are strategically combined to make ensemble techniques' output robust. The base detectors could consist of different kinds of learning algorithms of any number depending on the problem and adopted strategy \cite{le2021anomaly, ganaie2021ensemble}. The base detectors adopted for all ensemble methods in this work are OCSVM, OCNN, and IF. OCSVM is an unsupervised anomaly detection technique that learns a decision function for separating the normal class from anomalies in a dataset. OCSVM has been employed for ICS anomaly detection in \cite{raghuraman2021detecting, zhang2019anomaly, aboah2022plc}. OCNN is an unsupervised machine learning algorithm for anomaly detection formulated on the foundation of OCSVM \cite{aboah2022plc}. OCNN has been employed for anomaly detection in \cite{chalapathy2018anomaly}, although its application in ICS is very limited \cite{boateng2021anomaly, aboah2022plc}. IF is also an unsupervised machine learning technique that builds an ensemble of binary trees for a given dataset for anomaly detection \cite{liu2008isolation, hariri2019extended}. IF has been utilized to successfully detect anomalies in ICS \cite{elnour2020hybrid, togbe2020anomaly, aboah2022plc}. The ensemble methods proposed in this work use different voting strategies for combining OCSVM, OCNN, and IF outputs.

\subsection{Ensemble-Based Detection Approach}
 Ensemble learning is a technique for combining the outputs of multiple algorithms to create an improved output. The three different anomaly detection techniques mentioned in the subsection above make independent assumptions about the nature of the dataset, which may not be the case for every dataset. Ensemble learning techniques have been successfully employed in various data mining and supervised machine learning applications such as classification and recommender systems \cite{sayadi2018ensemble, dong2020survey, forouzandeh2021presentation}. However, unsupervised ensemble analysis is a recent field in anomaly detection, especially in ICS, with little work on combining anomaly detection algorithms \cite{aggarwal2017introduction}. This work seeks to extend the knowledge of ensemble-based detection approaches for anomaly detection in ICS. This section discusses five proposed ensemble-based anomaly detection techniques that combine the outputs of the three base detectors, namely, OCSVM, OCNN, and IF.

\begin{figure}[!t]
\centering
\includegraphics[width=9 cm]{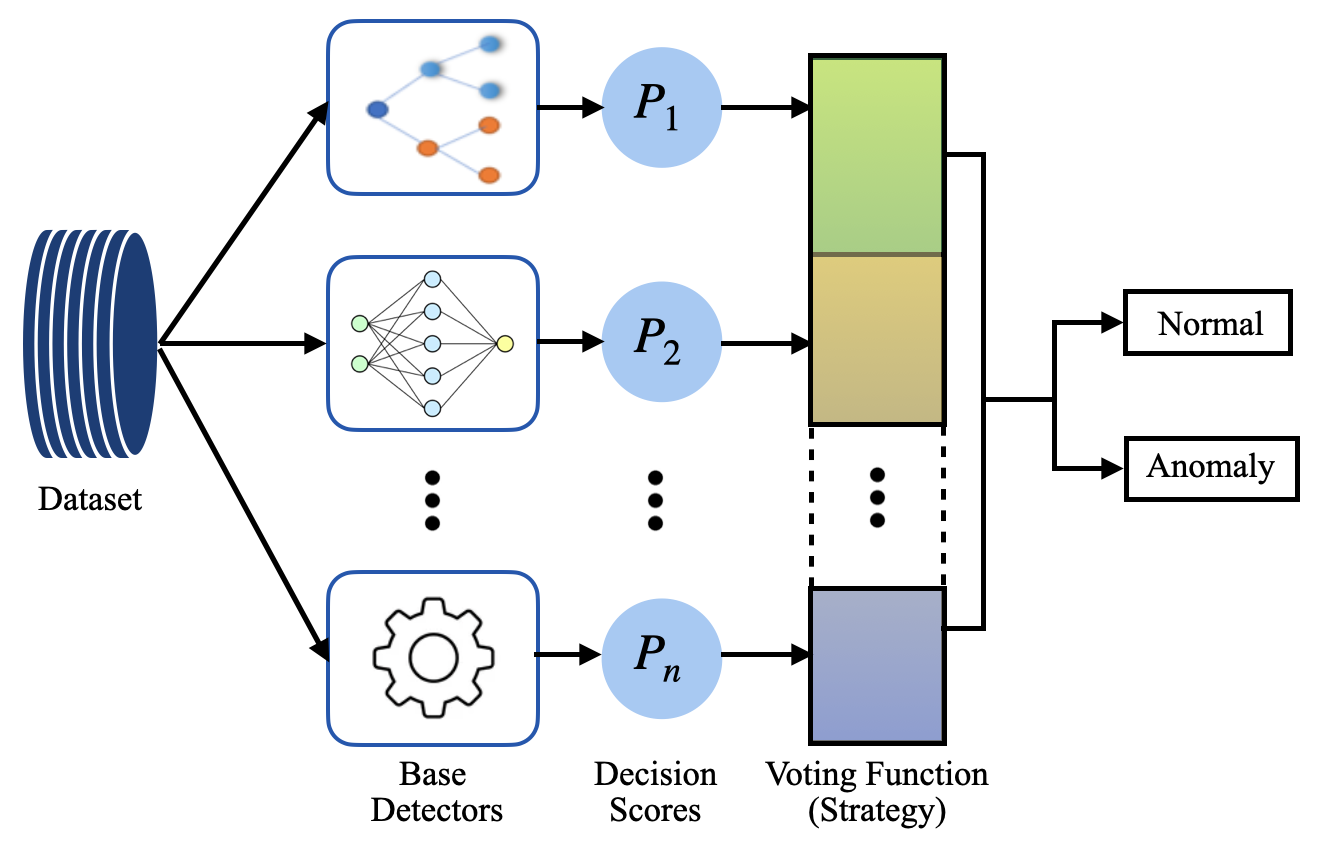}
\caption{General Representation of Ensemble Approach}
\label{fig2}
\end{figure}

Ensemble learning is generally challenging in the context of anomaly detection because of the unsupervised nature or labeled data unavailability \cite{aggarwal2017introduction}. Another challenge is the coherent combination of base detectors' decision scores, as base detectors may have different objective and scoring functions. Figure~\ref{fig2} shows a general representation of the proposed ensemble approach. Algorithm \ref{alg:one} and Algorithm \ref{alg:two} summarize the general approach to training and testing the proposed ensemble methods in this work. The major changes in each method relate to combining the scores of the base detectors and making final predictions, as described in the following subsections.

 \begin{algorithm}
 \caption{Anomaly detection ensemble algorithm training}
 \label{alg:one}
 \begin{algorithmic}[1]
 \renewcommand{\algorithmicrequire}{\textbf{Input:}}
 \renewcommand{\algorithmicensure}{\textbf{Output:}}
 \REQUIRE Input dataset $\{X_i|i=1,2,3...,n\}$
 \ENSURE  Decision score $S_i$ 
 \\ \textit{Training Phase} :
  \FOR {($k$ algorithm in base detectors)}
    \STATE Let $L_k$ be empty list of base detector decision scores
    \FOR{(data point $X_i$ in training dataset $X$)}
       \STATE Make prediction (anomaly score $S_{k}(i)$)
       \STATE Append decision score to the list: $L_k \gets S_{k}(i)$
    \ENDFOR
    \STATE {Normalize both positive and negative scores: \\ $S_{k}^{'}(i)$ = $\frac{L_{k_i} - \min{(L_{k})}}{\max{(L_k)} - \min{(L_k)}}$}
    \STATE {Store normalization feature range: \\ $F_{k_{norm}} = \min{(L_{k})}, \max{(L_k)}$}
  \ENDFOR
  \STATE Concatenate normalized scores $S_{k}^{'}(i)$ of base detectors as new multidimensional dataset, $D$ with $d_k(i) \in D$
  \STATE Let $S_i$ be the final ensemble prediction of data point $i$
  \FOR {$i$-th data point $d_k(i)$ in $D$}
    \STATE Make final prediction \\ \small{\texttt{/* Voting function is method specific */}}
  \ENDFOR
  \IF{$(S_i \geq 0)$}
    \STATE $X_i$ is normal instance
  \ELSE
    \STATE $X_i$ is anomalous instance
  \ENDIF
 \end{algorithmic} 
 \end{algorithm}

 \begin{algorithm}
 \caption{Anomaly detection ensemble algorithm testing}
 \label{alg:two}
 \begin{algorithmic}[1]
 \renewcommand{\algorithmicrequire}{\textbf{Input:}}
 \renewcommand{\algorithmicensure}{\textbf{Output:}}
 \REQUIRE New data point $X$
 \ENSURE  Decision score $S_t$
 \\ \textit{Testing Phase} : prediction on new data points
  \FOR {($k$ algorithm in base detectors)}
     \STATE Make prediction (anomaly score $S_{k}(T)$) for $T$
     \STATE Normalize $S_{k}(T)$ using $F_{k_{norm}}$
     \STATE $S_{k}^{'}(T) \gets$ norm($S_{k}(T)$)
  \ENDFOR
  \STATE Concatenate normalized scores of base detectors 
  \STATE Make final test prediction \\  \small{\texttt{/* Voting function is method specific */}}
  \IF{$(S_t \geq 0)$}
    \STATE $T$ is normal instance
  \ELSE
    \STATE $T$ is anomalous instance
  \ENDIF
 \end{algorithmic} 
 \end{algorithm}

\subsubsection{Majority-Vote Ensemble Method}
This hard voting strategy is a majority-vote ensemble that selects the mode of the base detectors' predictions. The majority vote ensemble techniques for supervised machine learning classifiers have been widely studied in the literature \cite{aggarwal2017introduction, brown2010good, geron2017hands}. The approach in this work adopts a similar strategy in \cite{aggarwal2017introduction} with an additional normalization mechanism for purposes of unsupervised anomaly detection. The base detectors provide decision scores for each data point in the training set during training. The decision scores are converted into binary predictions of either normal or anomaly based on the scoring function in \eqref{eq1}. Ultimately, the majority prediction by the base detectors becomes the final output of the observations in the training set. Similarly, during testing, each trained base detector predicts the new observation. The mode of the predictions becomes the new observation's final class (either normal or anomaly). This method follows Algorithm \ref{alg:one} and Algorithm \ref{alg:two} for training and testing by using the following voting function for $S_i$, where $S_i$ is the final ensemble prediction and $S_{k}^{'}(i)$ is the normalized positive and negative decision scores

\begin{equation}
\label{eq1}
   S_i \gets \textrm{mode}(S_{k}^{'}(i)). 
\end{equation} 

The majority vote ensemble detection method often achieves a higher anomaly detection rate than the best base detector in the ensemble \cite{aggarwal2017introduction}. In situations where the base detectors are weak detectors, this ensemble technique can still provide good performance \cite{aggarwal2017introduction}. The majority vote ensemble method is a naive approach with several challenges. In some cases where the majority of the base detectors have related objective functions, there is a high likelihood that this ensemble technique will make biased decisions. This ensemble technique's high bias in prediction is evident in results achieved in this work, as discussed in section 5. In order to minimize the majority vote ensemble's bias effect, the next subsection introduces an ensemble anomaly detection technique based on the maximum score over all base detectors. 

\subsubsection{Maximum-score Ensemble Method}
In the case of the maximum-score ensemble method, the final anomaly score of a data point is its maximum decision score over all base detectors. Lal et al. \cite{lal2019orfdetector} employed a maximum-score based ensemble method for online fraud detection. However, in this work, the maximum decision score over all base detectors is

\begin{equation}
\label{eq2}
   \textrm{Maximum}(i) = \max_{k=1}^{n}S_{k}(i)
\end{equation}
where $S_k$ is the anomaly score of a base detector $k$. During training, the scores of each base detector are normalized between the range of $[0, 1]$ and the normalization parameters (feature range) are saved. The base detector with the highest normalized score gets to decide the final anomaly score of the ensemble based on \eqref{eq2}. During testing phase, the base detectors predictions on new data points are normalized based on the training data normalization feature range. Subsequently, the base detector with the highest normalized score becomes the ensemble detector's final anomaly score. Maximum-score ensemble approach follows Algorithm \ref{alg:one} and Algorithm \ref{alg:two} for training and testing by using a voting function of \eqref{eq2}.

The effect of maximum-score ensemble learning is more complex because it can often improve bias but increase variance. This challenge makes the overall effect unpredictable \cite{aggarwal2017introduction}. In order to combat the high variance effect of this technique, the next section introduces a soft voting ensemble technique for anomaly detection.

\subsubsection{Soft Voting Ensemble Anomaly Detection Technique}
The soft voting ensemble anomaly detection technique reports the average decision scores of all base detectors as data points final scores. Soft voting ensemble methods based on scores averaging have been widely studied in the literature \cite{aggarwal2017introduction, geron2017hands, fathi2019improving}. The approach adopted in this work follows a similar technique in \cite{aggarwal2017introduction}, in order to compare its performance to other novel unsupervised methods proposed in this work. In this work, positive and negative decision scores of the base detectors are normalized during training, and the anomaly score of the $ith$ data point is computed as

\begin{equation}
\label{eq3}
   \textrm{Average}(i) = \frac{\displaystyle\sum_{k=1}^{n}S_k(i)}{n}
\end{equation}

During testing phase, base detectors' decision scores are normalized using the maximum and minimum normalization feature values of the training dataset, and the average is computed as the final score. If the average score is negative, the data point is an anomaly; otherwise, the data point is a normal instant. The soft voting ensemble approach follows Algorithm \ref{alg:one} and Algorithm \ref{alg:two} for training and testing by using a voting function of \eqref{eq3}. The intent of averaging the normalized scores is to reduce the variance of the scores and thereby improve accuracy.

\subsubsection{Weighted Voting Ensemble Technique}
The proposed weighted voting technique is an ensemble technique that produces a combined output prediction based on weights applied to the scores of the base detectors. The challenge of the soft voting technique for combining base detectors outputs is the appropriate weight assignment to each base detector. The work in \cite{aggarwal2017introduction,paulheim2015decomposition} show that uncorrelated features with all other features in a dataset are less relevant. Hence, uncorrelated scores violate the model of normal data dependencies. This work introduces a new generalized method of assigning weights to the base detectors based on statistical coefficient of determination (R-squared).  

Each base detector is trained on the training set to produce an anomaly score for each data point in the training set. The decision scores for each base detector are normalized according to Algorithm \ref{alg:one}, and concatenated to form features of a new training set. In this approach, the voting function in Algorithm \ref{alg:one} is replaced by a linear regression model. The linear regression model is used to predict each of the features from the other features and computes the root-mean-squared error $RMSE_{f_i}$ of predicting the $f_i$ feature from the other features. The weighted voting ensemble method can take an arbitrary learning model other than linear regression model such as decision tree regressor \cite{rokach2005decision}, K-nearest neighbor \cite{peterson2009k} and ridge regression \cite{mcdonald2009ridge}.  Irrelevant features will produce larger $RMSE_{f_i}$ \cite{aggarwal2017introduction, paulheim2015decomposition}, and hence larger $RMSE_{f_i}$ leads to lower associated feature weight. Next, each feature weight $f_i$ of the new training set is given by $ W_{f_i} = \max{(0, 1-RMSE_{f_i})}$. The weights $W_{f_i}$ of each feature are multiplied by the respective feature set $f_i$ to produce a weighted score for each feature. Finally, the maximum of the weighted features becomes the final anomaly score. Similarly, during prediction on a new data point after training, outputs of the base detectors are normalized and multiplied by the weights, $W_{f_i}$ obtained during training before the maximum score is selected as the new instance's final decision score.

The challenge with using $RMSE_{f_i}$ is that it works on the assumption that predicted scores of each base detector are normalized, making it difficult to scale or generalize to any arbitrary base detector. This work further proposes the use of R-squared (coefficient of determination) value $R^2$, as the weight of each feature $f_i$ instead of $\max{(0, 1-RMSE_{f_i})}$. The $R^2$-value  measures the level of explainability of predicting each feature from other features. It is easy and intuitive to calculate the $R^2$-value, and most importantly, it produces result in the range between 0 and 1 regardless of the input data scale. Higher $R^2$-value is better. $R^2$-value has a direct relationship with $MSE_{f_i}$ as

\begin{equation*}
\label{eq4}
    R^{2}_{f_k} = 1 - \frac{MSE_{f_k}}{Var(y)}
\end{equation*}
where
\begin{equation*}
    MSE_{f_k} = \frac{1}{N} \displaystyle\sum_{n=1}^{N} (y_i - f(x_i))^2
\end{equation*}
and 
\begin{equation*}
    \textrm{var}{(y)} = \displaystyle\frac{1}{N} \sum_{n=1}^{N} (y_i - E[y])^2
\end{equation*}
where $\textrm{var}(y)$ represents the total sum of squares (outcome response variance), $y_i$ is the target (ground truth scores), $x_i$ represents the dependent variables which are the features in this case, and $E[y]$ is average of $y$. The proposed approach for assigning weights to the output scores of the base detectors is summarized in Algorithm \ref{alg:three}. Substituting the voting function in Algorithm \ref{alg:one} with Algorithm \ref{alg:three} provides a complete Algorithm for the proposed weighted voting ensemble technique. The proposed weighted voting ensemble technique is a flexible algorithm because, instead of using the maximum of the weighted features as the final anomaly score in this work, the average of the weighted features could be used as the final anomaly score.

 \begin{algorithm}
 \caption{WV ensemble technique voting function}
 \label{alg:three}
 \begin{algorithmic}[1]
 \renewcommand{\algorithmicrequire}{\textbf{Input:}}
 \renewcommand{\algorithmicensure}{\textbf{Output:}}
 \REQUIRE  Dataset $D$, with $d_k(i) \in D$
 \ENSURE  Decision score $S_i$
 \\ \textit{Testing Phase} : prediction on new data points
  \FOR {(each feature $L_k$ in $D$)}
     \STATE predict $L_{k_i}$ from other features using linear regression
     \STATE $W_{L_k} \gets R^{2}_{L_k}$
  \ENDFOR
  \STATE $S_i \gets \max(W_{L_1} \times L_{1}, W_{L_2} \times L_{2}, \cdots, W_{L_k} \times L_{k})$
 \end{algorithmic} 
 \end{algorithm}

\subsubsection{Stacking-based Anomaly Detection Approach}
This work proposes a new stacking-based anomaly detection architecture to identify ICS anomalies. The approach adopted here uses a meta-detector, often called blender, to best learn how to combine the output scores of the base detectors to produce robust, stable, and accurate results. First, each base detector is trained independently on the training dataset. The base detectors have different scoring functions; hence their outputs are normalized as described in Algorithm \ref{alg:one}. The normalized outputs of the base detectors are used as input (training data) to the meta-detector to learn the best way to combine the normalized scores to produce a robust anomaly detection model. 

\begin{figure}[!t]
\centering
\includegraphics[width=8 cm]{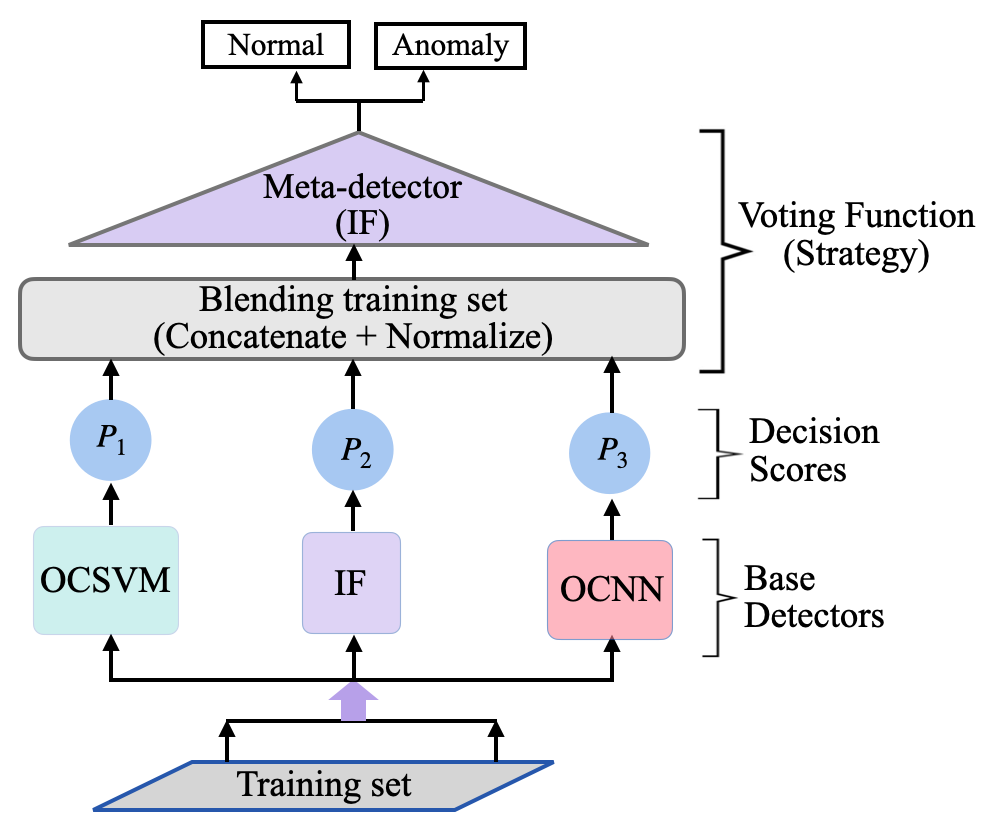}
\caption{Stacking-based anomaly detection architecture}
\label{fig3}
\end{figure}

IF is used as the meta-detector for learning the new training set because the detection problem is still unsupervised, so labeled classes of the training data are assumed to be unavailable. Again, IF is a robust and fast anomaly detection algorithm that builds a homogeneous binary tree ensemble for a given dataset \cite{aboah2022plc}. Figure~\ref{fig3} shows the architecture of the proposed stacking-based anomaly detection approach. Replacing the voting function in Algorithm \ref{alg:one} with a layer of IF meta-detector described in this section results in the complete algorithm for this proposed technique. Each base detector provides an independent normalized score for the new data point during model prediction after training, which serves as input to the meta-detector. Finally, the meta-detector provides a combined score for the new data point to determine whether it is a normal data point or an anomaly. Table~\ref{tab2} shows optimal hyperparameters used to develop the IF meta-detector. 

\begin{table}
\begin{center}
\caption{Model hyperparameters for IF meta-detector in Figure~\ref{fig3}}
\label{tab2}
\begin{tabular}{| c | p{4.5cm} | c |}
\hline
\textbf{Parameter}	& \textbf{Description}	& \textbf{value} \\
\hline
$n_{estimators}$      & Number of base estimators in the forest ensemble & 100\\
\hline
$n_{max}$		& Number of training samples to draw to train each estimator	& 256\\
\hline
contamination	& Proportion of outliers in the data set	& 0.007\\
\hline
\end{tabular}
\end{center}
\end{table}

\section{Results and Discussions}
The evaluation is based on performance metrics, results from predictions on the test data, and comparison with prior work trained on a similar dataset. Google's Tensorflow \cite{tensorflow2015-whitepaper}, an open-source deep learning library, is used for training and performing inference on the base detectors. Evaluation results and performance metrics calculations are performed by using the Scikit-learn library \cite{scikit-learn}. The dataset used to develop the proposed methods in this work is the HMI historian log of operations at PLC memory addresses. The data is obtained through Modbus communication protocol between the PLC and HMI.

The performance metrics for evaluating the proposed anomaly detection models are derived from the confusion matrix. Four evaluation outcomes are derived from the confusion matrix: true positive (TN), true negative (TN), false positive (FP), and false negative (FN). These outcomes are used for calculating the accuracy, precision, recall, and F1-score of anomaly detection models as described in \cite{aboah2022plc}.

The results presented in this section use the visualization approach proposed in \cite{aboah2022plc, boateng2022new} to better understand ensemble algorithms' performance. The histogram-based visualization approach normalizes the histogram frequency (y-axis) to a range between $0\%$ and $100\%$, whereas the x-axis represents the normalized decision scores indicating the prediction confidence. After investigating various hyperparameter ranges, the three base detectors, OSCVM, OCNN, and IF, are trained with optimal hyperparameters. The optimal hyperparameters for the base detectors are presented in \cite{aboah2022plc}.

\subsection{Performance of Majority-Vote Ensemble Method}
\begin{figure}[!t]
\centering
\includegraphics[width=9 cm]{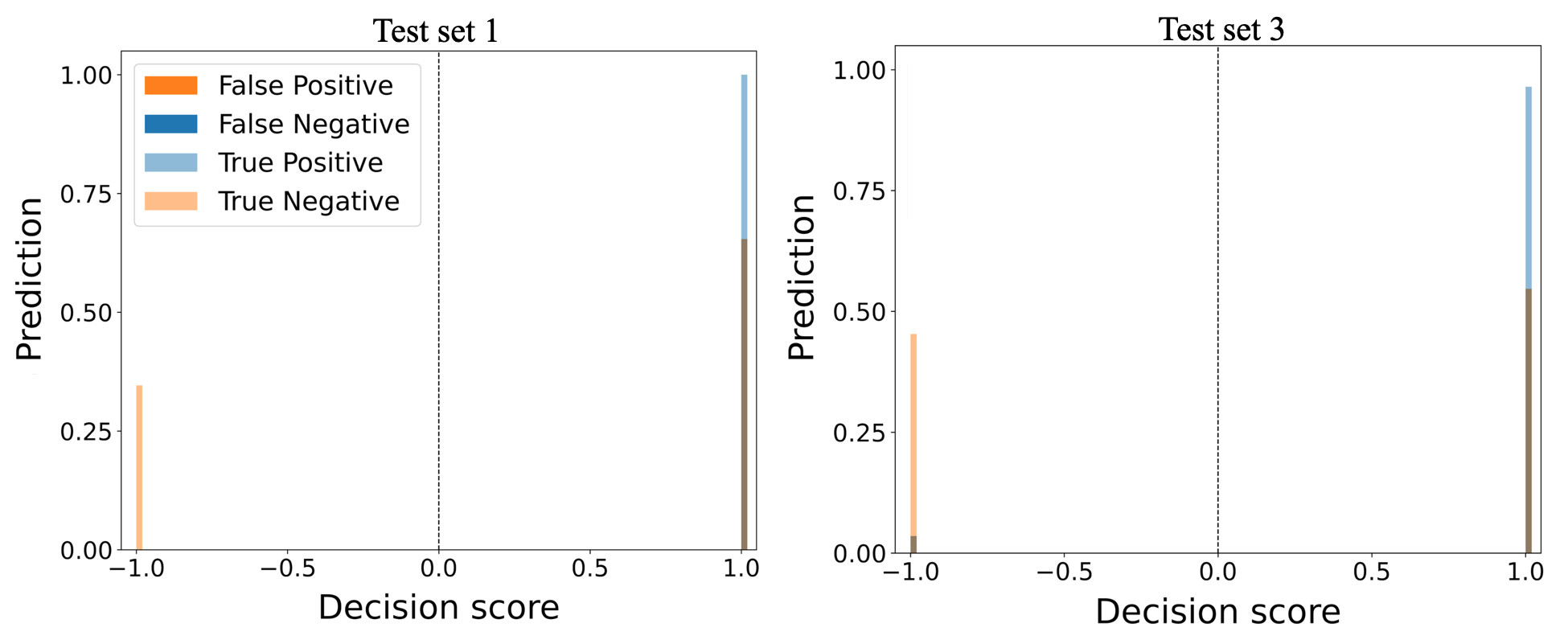}
\caption{Majority-vote ensemble method results of the normalized TP, TN, FP, and FN values on test sets 1 and 3}
\label{fig4}
\end{figure}

The majority-vote ensemble method performed well on test set 1 by having precision, recall, and F1-score of $90\%$, $91\%$, and $90\%$ respectively. The algorithm achieved a recall and F1-score of $88\%$ and $84\%$ respectively on test set 2, which are similar to the recall and F1-score values of test set 3. However, the algorithm's high recall and F1-score rate on test set 2 resulted from its high anomaly detection rate on the normal data points at the expense of high false positive rate. The algorithm finds it challenging to detect anomalies involving anomalous scenarios 1 and 3. Performance of the algorithm decreases on test sets 4 and 5 as the recall values decreased to $82\%$ and $71\%$ respectively. Figure~\ref{fig4} shows the majority-vote ensemble method results of the normalized TP, TN, FP, and FN values on test sets 1 and 3. 

Because of the algorithm's binary prediction outcome, it is impossible to visualize and assess prediction confidence levels, unlike other ensemble approaches proposed in this work. Figure~\ref{fig4} shows that although the algorithm's TN predictions are less than $50\%$, the levels of FN predictions are low at values below $10\%$. However, the algorithm has high levels of FP in the range above $50\%$ for test sets 1 and 2. The visualization of the algorithm performance on test sets 3-5 is similar to Figure~\ref{fig4}.

\subsection{Performance of Maximum-score Ensemble Method}
\begin{figure}[!t]
\centering
\includegraphics[width=9 cm]{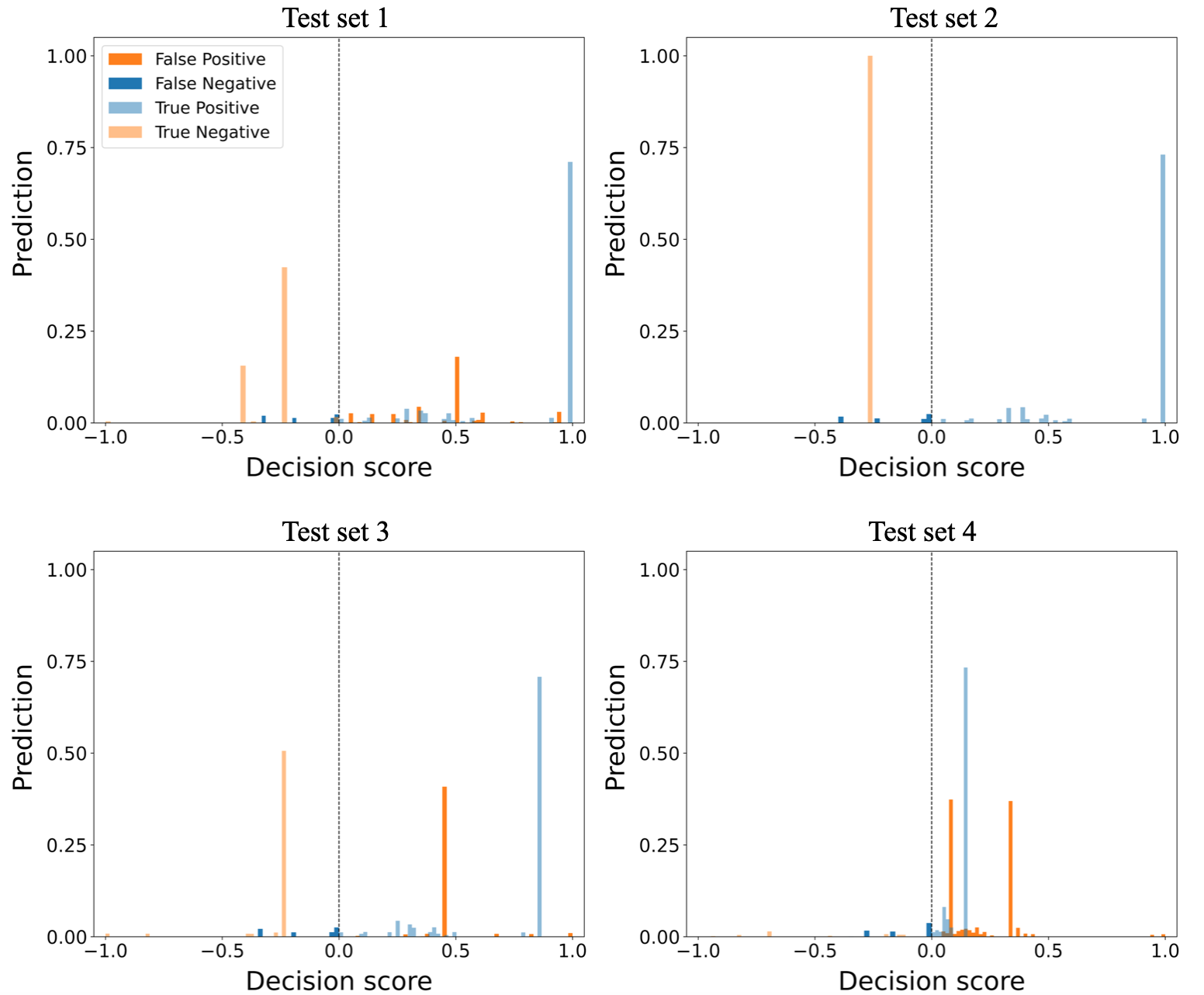}
\caption{Maximum-score ensemble method results of the normalized TP, TN, FP, and FN values on test sets 1-4}
\label{fig5}
\end{figure}

Maximum-score ensemble method performance on test sets 1 and 2 are high with F1-scores of $90\%$ and $95\%$ respectively. This method has the same precision, and recall values as the majority-vote method on test set 1. However, this method significantly performs well on test set 2 as compared to the majority-vote method by having precision and recall values of $96\%$ and $94\%$ respectively. The algorithm's performance on test sets 3 and 5 is similar to majority-vote method. F1-scores for test sets 3 and 5 are $85\%$ and $71\%$, respectively. This method has its worst performance on test set 4, with recall and F1-scores of $66\%$ and $58\%$ respectively. In effect, the maximum-score ensemble method has low detection performance on TLIGHT system anomalies involving timing bits manipulation. Figure~\ref{fig5} shows the maximum-score ensemble method results of the normalized TP, TN, FP, and FN values on test sets 1-4. 

Maximum-score ensemble method has TP values above $60\%$ in test sets 1-4 and high TP scores confidence in test sets 1-3. The method detected all anomalies in test set 2 by having a TN value of $100\%$, which explains its high performance on test set 2. Figure~\ref{fig5} shows that the method misclassified over $95\%$ of test set 4 anomalies as normal instances, which resulted in high FP and low TN values. The visualization of the algorithm performance on test set 5 is similar to test set 4, and therefore it is not shown here.

\subsection{Performance of Soft Voting Ensemble Method}
\begin{figure}[!t]
\centering
\includegraphics[width=9 cm]{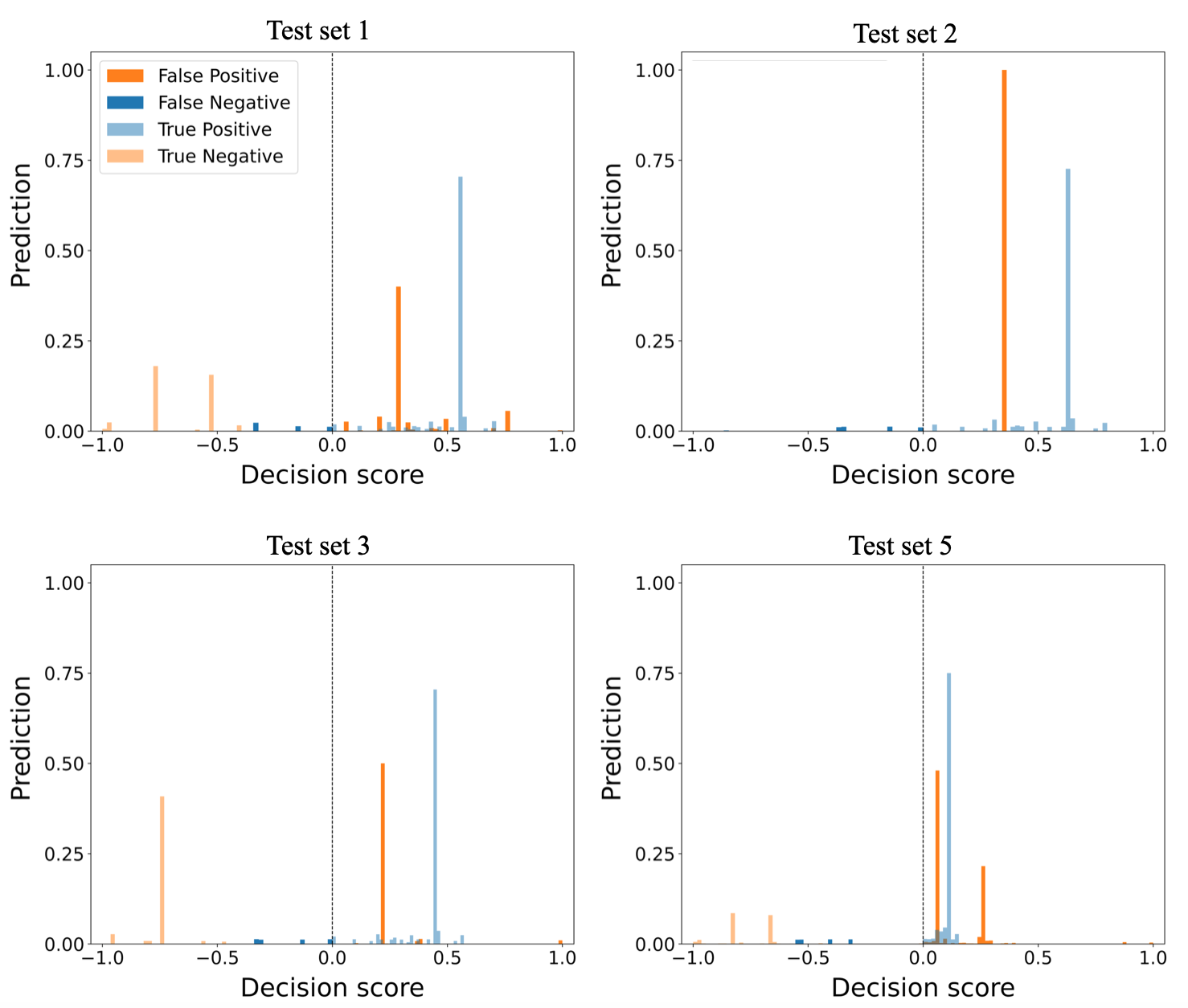}
\caption{Soft voting ensemble method results of the normalized TP, TN, FP, and FN values on test sets 1, 2, 3, and 5}
\label{fig6}
\end{figure}

Soft voting ensemble method achieved high performance on test set 1 with a precision, recall, and F1-score of $89\%$, $90\%$, and $89\%$ respectively. The method's performance on test sets 1 and 2 are similar, with F1-scores of $83\%$ and $84\%$ respectively. However, the method has low performance on test sets 4 and 5, with F1-scores in both cases below $60\%$. The method's low performance on test sets 4 and 5 signifies its inability to detect timing bits anomalies. Figure~\ref{fig6} shows the soft voting ensemble method results of the normalized TP, TN, FP, and FN values on test sets 1, 2, 3, and 5. 

Soft voting ensemble method has a TP prediction confidence over $50\%$ on test sets 1 and 2. Again, the method correctly predicted the TLIGHT system's normal behavior in all test cases by having TP predictions over $60\%$. However, the method misclassified all anomalies in test set 2 as normal instances leading to a high FP rate of $100\%$, which explains the low performance of the algorithm on test set 2. The method's confidence levels of TP and FP predictions on test set 5 are below $50\%$. Also, over $80\%$ of anomalies in test set 5 are misclassified as normal instances leading to high FP rate. The visualization of the algorithm performance on test set 4 is similar to test set 5, and therefore it is not shown here.

\subsection{Performance of Weighted Voting Ensemble Method}
\begin{figure}[!t]
\centering
\includegraphics[width=9 cm]{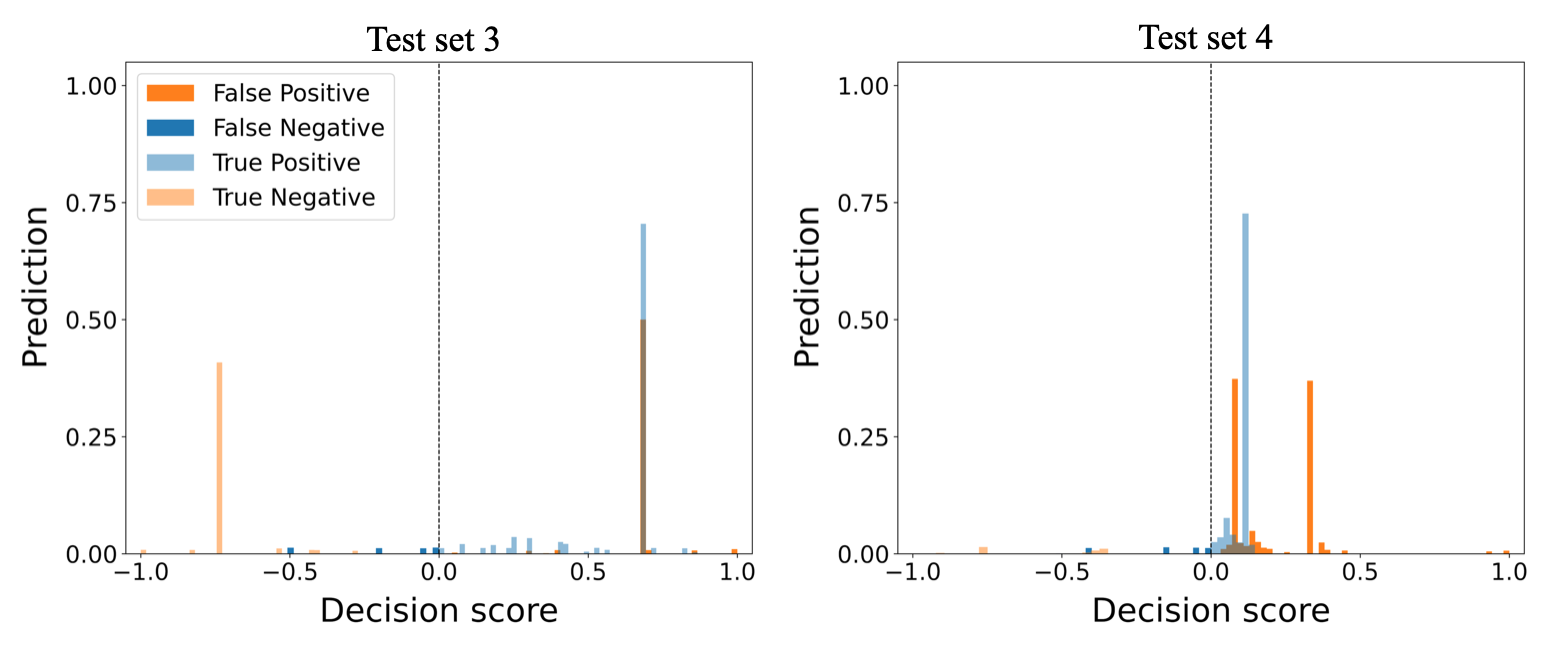}
\caption{WV-OLS ensemble method results of the normalized TP, TN, FP, and FN values on test sets 3 and 4}
\label{fig7}
\end{figure}

The weighted voting ensemble method uses an ordinary least squares regression model for assigning weights to the base detectors. The weighted voting with ordinary least squares regression model WV-OLS has a similar performance to soft voting ensemble method on all datasets. The method achieved its highest performance on test set 1 with precision, recall, and F1-score of $89\%$. The method's performance on test set 2 and 3 are similar, with F1-scores of $83\%$ and $84\%$ respectively. This method has its lowest performance on test sets 4 and 5, with F1-scores below $60\%$. The method's low performance on test sets 4 and 5 signifies its inability to detect anomalies in timing bits. Figure~\ref{fig7} shows the WV-OLS ensemble method results of the normalized TP, TN, FP, and FN values on test sets 3 and 4.

The WV-OLS method correctly predicted the TLIGHT system's normal behavior in all test sets with a TP score over 60\%. The method's confidence of TP and FP predictions on test set 3 is over $50\%$, whereas the confidence of the TP and FP predictions on test set 4 is below $50\%$. The visualization of the algorithm performance on test sets 1 and 2 are similar to test set 3, whereas the visualization of the algorithm performance on test sets 5 is similar to test set 4, and therefore they are not shown here. The method misclassified over $80\%$ of anomalies as normal in test set 4 leading to high FP. The method's high FP value on test set 4 explains its low performance on test set 4.

\begin{figure}[!t]
\centering
\includegraphics[width=9 cm]{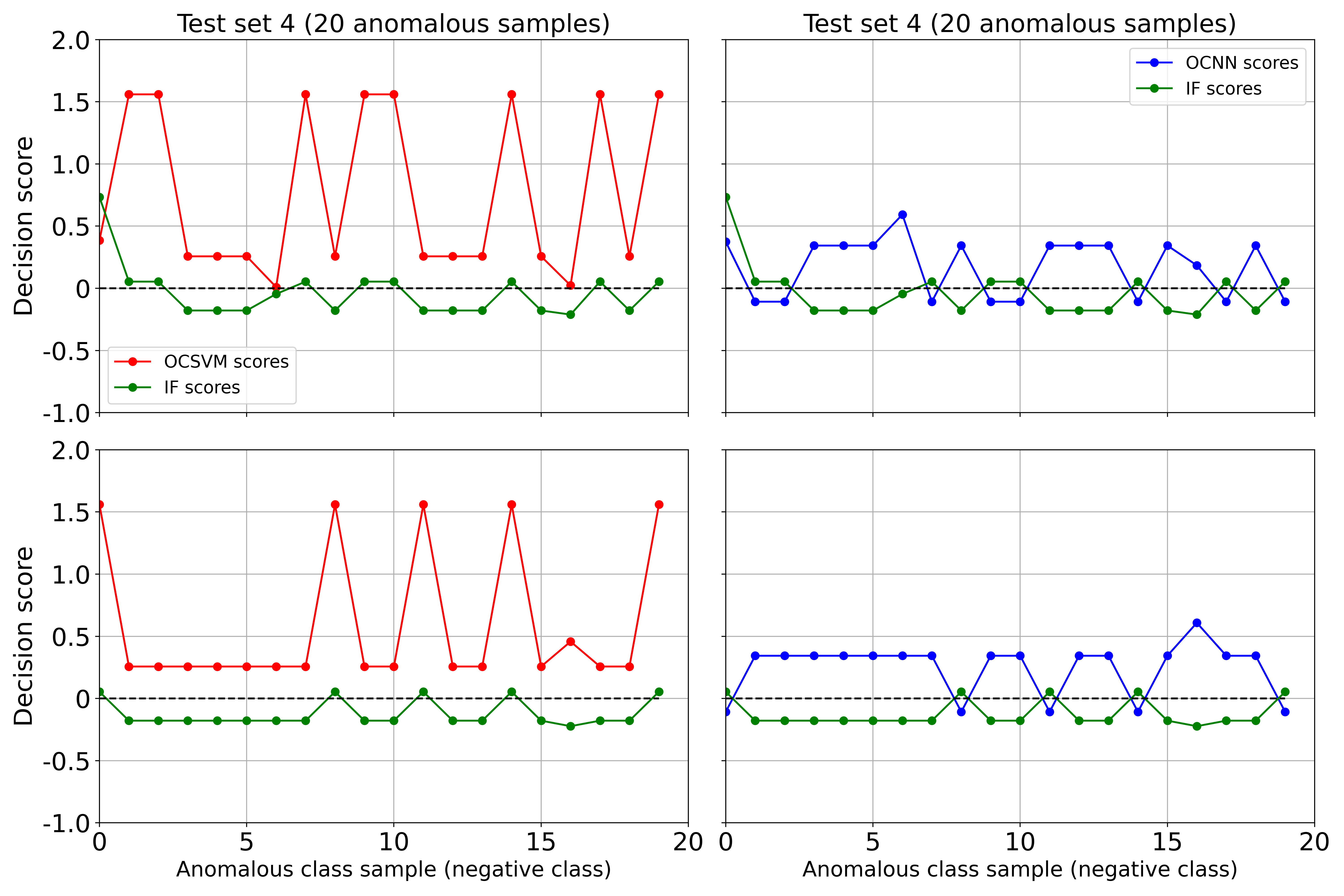}
\caption{WV-OLS base detectors predictions on 20 anomalous samples of test set 4}
\label{fig8}
\end{figure}

WV-OLS ensemble method performance is lower than maximum score ensemble method. The WV-OLS ensemble method's low performance results from the similarities between OCSVM and OCNN objective functions which makes their errors correlated. OCNN is formulated based on the foundation of OCSVM optimization problem \cite{aboah2022plc}; hence WV-OLS method's assigned weights are biased towards OCSVM and OCNN. Figure~\ref{fig8} shows the decision scores of WV-OLS base dectectors on 20 anomalous samples of test set 4. In each of the 20 anomalous samples, a negative score is desired to signify correct prediction. However, the OCSVM base detector misclassifies the anomalous test samples by producing positive scores for all 20 anomalous test samples. The OCNN base detector misclassifies more than 50\% of the anomalous test samples. On the contrary, the IF base detector correctly detects more than 60\% of the anomalous test samples. Similar misclassifications of anomalous test samples by the OCSVM and OCNN base detectors were observed across all test sets, explaining the low performance of the WV-OLS approach to weighted voting.

In an attempt to control the WV-OLS method's high bias, the ordinary least squares linear regression model for assigning weights in Algorithm \ref{alg:three} can be replaced by a ridge regression model \cite{rokach2005decision} or K-nearest neighbor model \cite{peterson2009k}. The resulting weighted voting approaches using these learning models are denoted WV-Ridge and WV-KNN, respectively. The proposed weighted voting method using different linear regression models are compared with the weighted voting approach found in \cite{aggarwal2017introduction}.  WV-\cite{aggarwal2017introduction}  uses RMSE for assigning weights to the base detectors. Table~\ref{tab3} shows accuracy, precision, recall, and F1-score for the four weighted voting approaches. The WV-OLS approach shows no significant improvement over WV-Ridge and weighted voting method in previous work \cite{aggarwal2017introduction}. WV-OLS, WV-Ridge, and WV-\cite{aggarwal2017introduction} all use linear models for assigning weights to base detectors' output; therefore, the same fundamental limitation associated with the bias of OCSVM and OCNN base detectors holds. However, WV-KNN outperforms WV-\cite{aggarwal2017introduction}, WV-OLS, and WV-Ridge on test sets 1, 2, and 5 across all performance metrics.  WV-KNN achieved over 7\% improvement in accuracy, precision, recall, and F1-score on test set 2 as compared to WV-\cite{aggarwal2017introduction}, WV-OLS, and WV-Ridge. WV-KNN achieved high performance because it relies on distance in the feature space instead of making a linear assumption about the dataset.

\begin{table}
\begin{center}
\caption{Performance comparison of weighted voting ensemble methods}
\label{tab3}
\begin{tabular}{| p{0.8cm} | p{1.3cm} | p{1.0cm} | p{1.0cm} | p{0.9cm} | p{1.1cm} |}
\hline
\textbf{Dataset} & \textbf{WV Method} & \textbf{Accuracy} & \textbf{Precision} & \textbf{Recall} & \textbf{F1-Score}\\
\hline
  Test & WV-\cite{aggarwal2017introduction} & 0.89  &0.88  &0.89  & 0.89\\
  Set & WV-OLS & 0.89  &0.89  &0.89  & 0.89\\
  1 & WV-Ridge & 0.89 &0.89  &0.89  & 0.89\\
   & WV-KNN & \textbf{0.90}  & \textbf{0.91}  & \textbf{0.90}  & \textbf{0.90}\\
   \hline
     Test & WV-\cite{aggarwal2017introduction} & 0.85  &0.81  &0.85  & 0.83\\
   Set & WV-OLS & 0.86 & 0.81  & 0.86  & 0.83\\
   2 & WV-Ridge & 0.86  &0.81  &0.86  & 0.83\\
   & WV-KNN & \textbf{0.94}  & \textbf{0.96}  & \textbf{0.94}  & \textbf{0.95}\\
   \hline
    Test & WV-\cite{aggarwal2017introduction} & 0.85  &0.83  &0.85  & 0.84\\
   Set & WV-OLS & 0.85  &0.84  &0.85  & 0.84\\
   3 & WV-Ridge & 0.85  &0.84  &0.85  & 0.84\\
   & WV-KNN & 0.85  &0.85  &0.85  & \textbf{0.85}\\
   \hline
     Test & WV-\cite{aggarwal2017introduction} & 0.67  &0.56  &0.67  & 0.58\\
   Set & WV-OLS & \textbf{0.67}  & \textbf{0.56}  & \textbf{0.67}  & 0.58\\
   4 & WV-Ridge & 0.67  &0.56  &0.67  & 0.58\\
   & WV-KNN & 0.66  &0.55  &0.66  & 0.58\\
   \hline
     Test & WV-\cite{aggarwal2017introduction} & 0.58  &0.67  &0.58  & 0.51\\
   Set & WV-OLS & 0.57  &0.67  &0.57  & 0.50\\
   5 & WV-Ridge & 0.57  &0.67  &0.57  & 0.50\\
   & WV-KNN & \textbf{0.73}  & \textbf{0.77}  & \textbf{0.73}  & \textbf{0.71}\\
\hline
\end{tabular}
\end{center}
\end{table}

\begin{figure}[!t]
\centering
\includegraphics[width=9 cm]{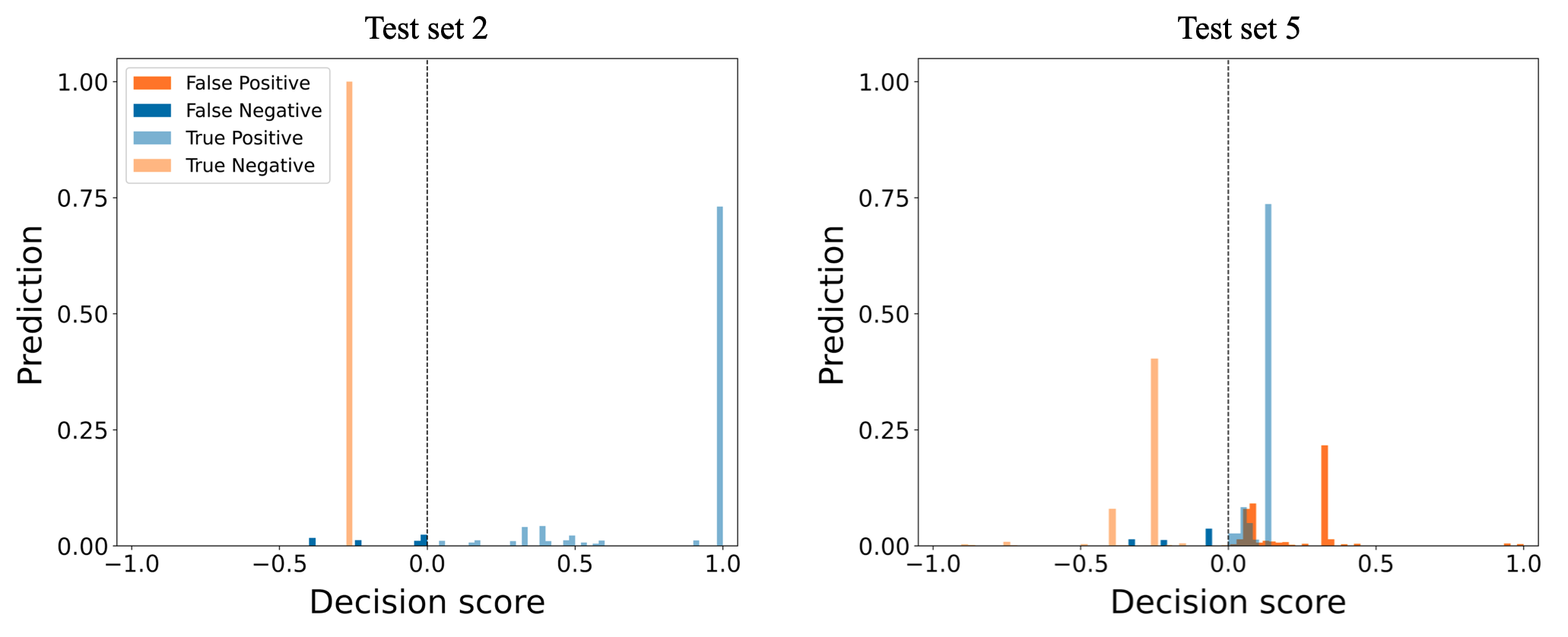}
\caption{WV-KNN ensemble method results of the normalized TP, TN, FP, and FN values on test sets 2 and 5}
\label{fig9}
\end{figure}

Figure~\ref{fig9} shows WV-KNN ensemble method results of the normalized TP, TN, FP, and FN values on test sets 2 and 5. WV-KNN correctly detects all anomalies in test set set with over 25\% prediction confidence. Also, WV-KNN correctly detects over 70\% of normal data points with 100\% prediction confidence leading to high TP value in test set 2. WV-KNN's high anomaly detection on test set 2 explains why it achieves over 90\% accuracy, precision, recall, and F1-score. WV-\cite{aggarwal2017introduction}, WV-OLS, and WV-Ridge are unable to detect more than 20\% of anomalies in test set 5. However, WV-KNN detects over 50\% of anomalies in test set 5 which explains why WV-KNN has over 10\% improved accuracy, precison, recall, and F1-score as compared to WV-\cite{aggarwal2017introduction}, WV-OLS, and WV-Ridge. The visualization of WV-KNN performance on test 1 and 3 are similar to WV-OLS visualization on test set 3 in Figure~\ref{fig7}. The visualization of WV-KNN performance on test set 4 is similar to the WV-OLS visualization on test set 4 in Figure~\ref{fig7}, and therefore it is not shown here. Lastly, WV-\cite{aggarwal2017introduction} and WV-Ridge performance visualizations across all test sets are similar to those of WV-OLS in Figure~\ref{fig7}, and therefore they are not shown here.

\subsection{Performance of Stacking-based Ensemble Method}
\begin{figure}[!t]
\centering
\includegraphics[width=9 cm]{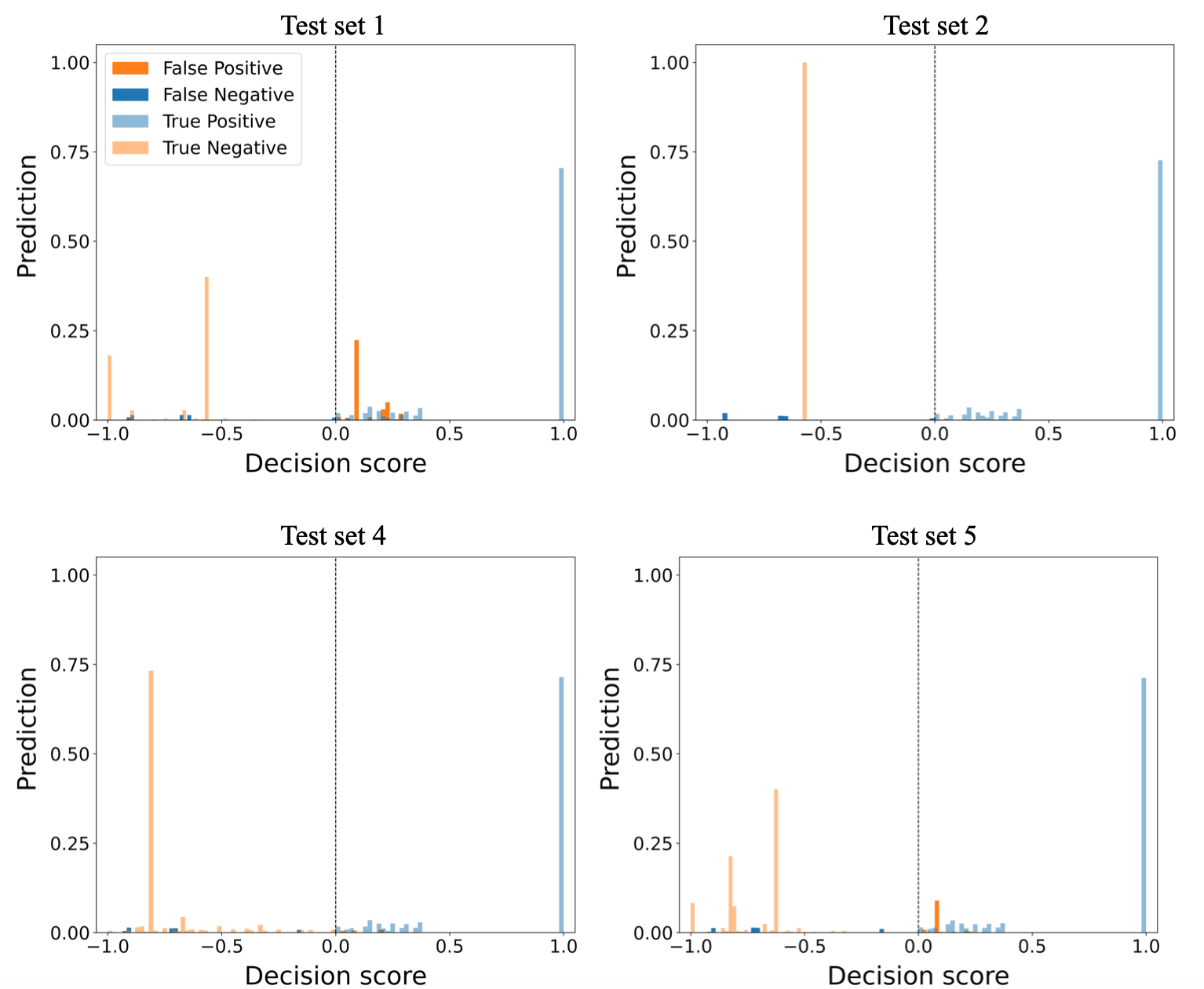}
\caption{Stacking-based ensemble method results of the normalized TP, TN, FP, and FN values on test sets 1, 2, 4, and 5}
\label{fig10}
\end{figure}

Stacking based ensemble method uses an IF meta-detector to consider the results of the base detectors.
This method has excellent performance on all test sets. It achieved its highest performance on test sets 2 and 4 by having the same precision, recall, and F1-score of $97\%$, $96\%$, and $96\%$, respectively. Unlike other ensemble methods described in previous sections, the stacking-based ensemble has high performance on test sets 4 and 5 with recall values of $96\%$ and $92\%$ respectively. The method's performance shows its ability to detect TLIGHT signals anomalies and its robustness in detecting timing bits anomalies. The method had its worst performance on test set 3 with a recall and F1-score of $87\%$ and $86\%$ respectively. The high performance of the stacking-based ensemble method is due to the IF meta-detector's ability to extract significant information from the base detectors in an unsupervised learning manner. Figure~\ref{fig10} shows the stacking-based ensemble method results of the normalized TP, TN, FP, and FN values on test sets 1, 2, 4, and 5.

This method correctly predicted all anomalies in test 2 with high prediction confidence and TN value of $100\%$, which explains why it attained an F1-score of $96\%$. This method correctly predicted TLIGHT system normal behavior with TP value over $70\%$ at $100\%$ prediction confidence in all test sets. Again, in all test sets, over $70\%$ of TN predictions have prediction confidence over $50\%$. A closer assessment of Figure~\ref{fig10} confirms that not only is the stacking-based method excellent at detecting anomalies, but in all test cases, the method is confident about its decisions. The visualization of the algorithm performance on test sets 3 is similar to test set 1, and therefore it is not shown here.

\begin{figure}[!t]
\centering
\includegraphics[width=9 cm]{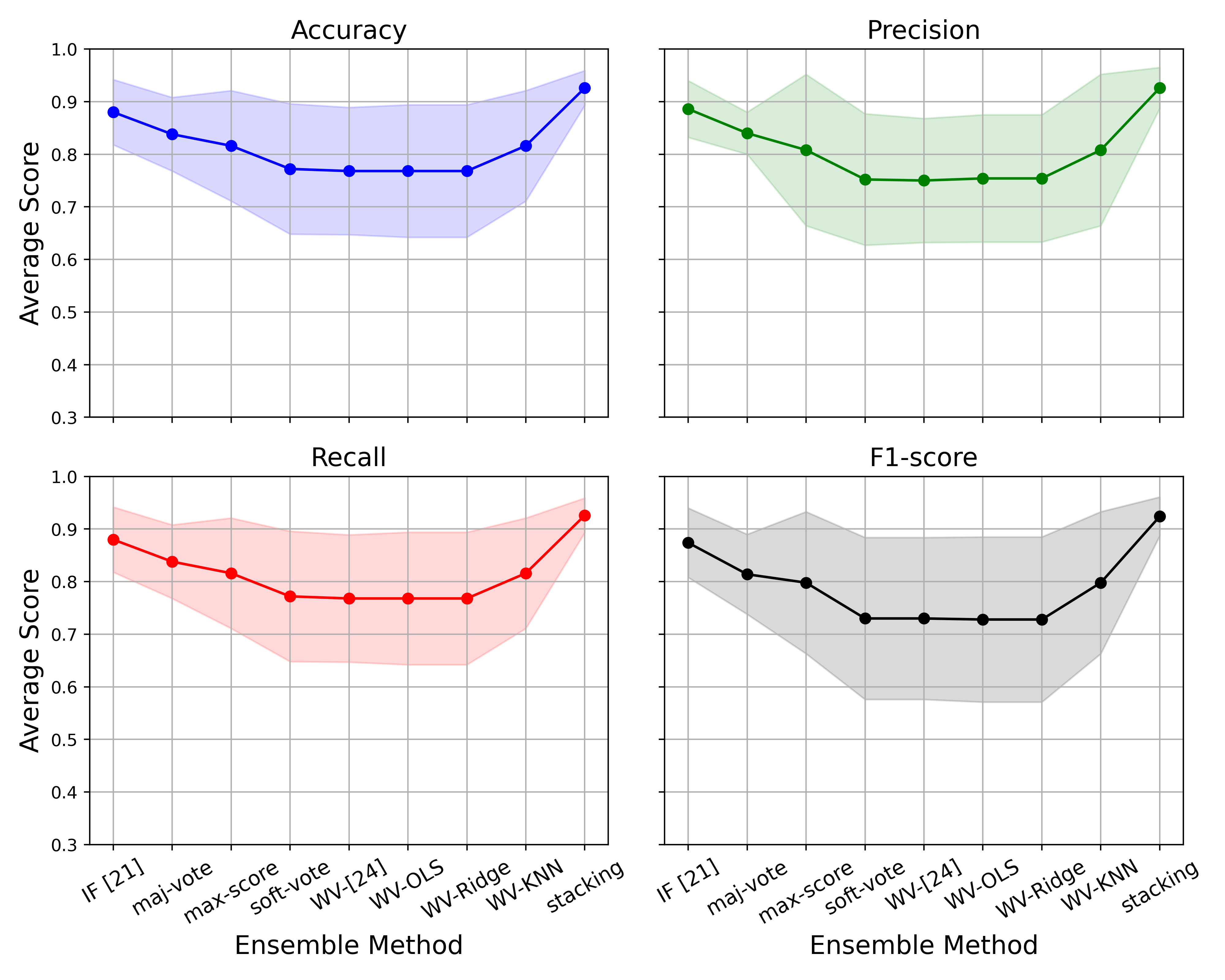}
\caption{Average performance of ensemble methods with their respective standard deviations}
\label{fig11}
\end{figure}

Figure~\ref{fig11} shows the average performance of the ensemble methods with their respective standard deviations on the five test sets. Stacking-based ensemble and IF \cite{aboah2022plc} have the least standard deviations $<$ 0.07 on all performance metrics across all test sets. WV \cite{aggarwal2014algorithms}, WV-OLS, WV-Ridge, and soft-voting methods have the high standard deviations $>$ 0.1 as a result of their poor performance on test sets 4 and 5. Table~\ref{tab4} provides a performance summary of all ensemble methods explored in this work. It is evident from Table~\ref{tab4} that stacking-based ensemble method has superior performance over all test sets. 

\begin{table*}
\begin{center}
\caption{Average performance summary of ensemble methods}
\label{tab4}
\begin{tabular}{| p{2.3cm} | p{0.8cm}  p{0.8cm} | p{0.8cm}  p{1cm} | p{0.8cm}  p{0.8cm} | p{1.2cm}  p{0.8cm} |}
\hline
 & \textbf{Accuracy}	& & \textbf{Precision} & & \textbf{Recall} & & \textbf{F1-score} &\\
 \textbf{Technique} & Mean & std & Mean & std & Mean & std & Mean & std \\
\hline
IF \cite{aboah2022plc}  &  0.880 & 0.062  & 0.886 & 0.054 & 0.880 & 0.062 & 0.874 & 0.066\\
\hline
Majority-vote & 0.838 & 0.070 & 0.840 & 0.040 & 0.838 & 0.070 & 0.814 & 0.076 \\
\hline
Maximum-score & 0.816 & 0.105 & 0.808 & 0.144 & 0.816 & 0.105 & 0.798 & 0.135 \\
\hline
Soft-voting & 0.772 & 0.124 & 0.752 & 0.125 & 0.772 & 0.124 & 0.730 & 0.154 \\
\hline
WV-\cite{aggarwal2017introduction} & 0.768& 0.121& 0.750& 0.118& 0.768& 0.121& 0.730 & 0.154\\
\hline
WV-OLS & 0.768 & 0.126 & 0.754 & 0.121 & 0.768 & 0.126 & 0.728 & 0.157 \\
\hline
WV-Ridge & 0.768 & 0.126 & 0.754 & 0.121 & 0.768 & 0.126 & 0.728 & 0.157 \\
\hline
WV-KNN & 0.816 & 0.105 & 0.808 & 0.144 & 0.816 & 0.105 & 0.798 & 0.135 \\
\hline
Stacking & \textbf{0.926} & 0.033 & \textbf{0.926} & 0.039 & \textbf{0.926} & 0.033 & \textbf{0.924} & 0.037 \\
\hline
\end{tabular}
\end{center}
\end{table*}

\subsection{Comparison with Previous Work}
An anomaly detection performance comparison between individual unsupervised machine learning detectors using the same dataset is presented in \cite{aboah2022plc}.  The unsupervised ensemble methods proposed here strategically aggregates the decision scores of the three methods examined in \cite{aboah2022plc}. This section compares the average performance of the proposed unsupervised ensemble techniques with the reported results of the highest performing algorithm in \cite{aboah2022plc} across all test sets.

\begin{figure}[!t]
\centering
\includegraphics[width=9 cm]{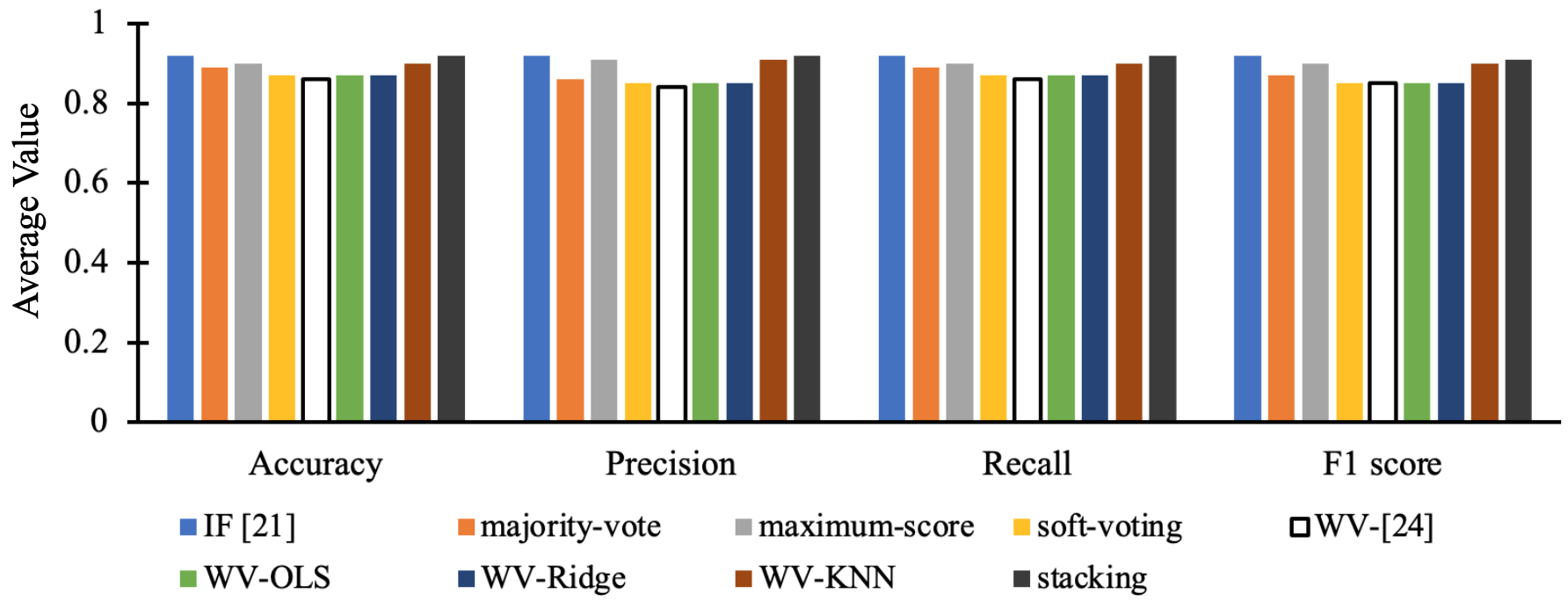}
\caption{Average test sets 1--3 performance of the proposed ensemble methods in this work and the reported results of IF in \cite{aboah2022plc}}
\label{fig12}
\end{figure}

First, the comparison is performed across test sets 1-3, which consisted of TLIGHT system signal anomalies, excluding timing bits anomalies. Figure~\ref{fig12} shows the average test sets 1--3 performance proposed ensemble methods in this work and the reported results of IF in \cite{aboah2022plc}. Soft-voting, WV-OLS, WV-Ridge and WV-\cite{aggarwal2017introduction} methods have similar average accuracy, precision, recall, and F1-score. Majority vote and WV-KNN have similar average accuracy, precision, recall, and F1-score. The stacking ensemble method and IF \cite{aboah2022plc} algorithm achieve the same average accuracy, precision, recall, and F1-score of $92\%$. Therefore, the stacking approach and IF \cite{aboah2022plc} have similar detection performance on TLIGHT system anomalies, excluding timing bit anomalies. Stacking ensemble and IF \cite{aboah2022plc} have superior performance to all other ensemble techniques.

\begin{figure}[!t]
\centering
\includegraphics[width=9 cm]{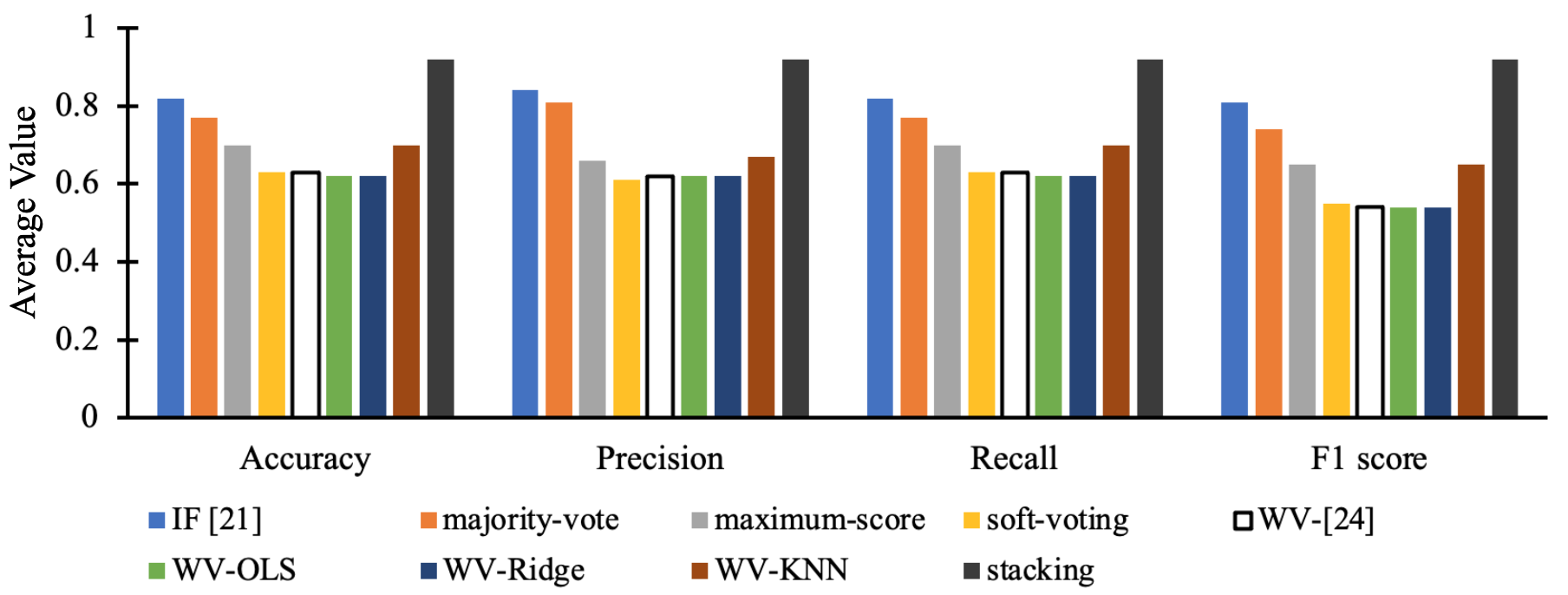}
\caption{Average test sets 4--5 performance of the proposed ensemble methods in this work and the reported results of IF in \cite{aboah2022plc}}
\label{fig13}
\end{figure}

Test sets 4--5 consist of timing bits anomalies unique to \cite{aboah2022plc} and this work. Figure~\ref{fig13} shows the average performance comparison between the ensemble methods presented in this work and the reported results of IF algorithm in \cite{aboah2022plc} on test sets 4--5. Soft-voting maximum-score, WV-\cite{aggarwal2017introduction},  WV-OLS, and WV-Ridge methods have relatively similar average accuracy, and precision on test sets 4--5 which were inferior to WV-KNN and maximum vote ensemble methods. Stacking-based ensemble achieves the same average accuracy, precision, recall, and F1-score of $92\%$, whereas IF \cite{aboah2022plc} achieves the same average accuracy, precision, recall, and F1-score of $82\%$. Overall, stacking-based ensemble method achieves outstanding performance on all evaluation metrics on test sets 4--5. The stacking-based ensemble method's high performance on all test sets compared to previous work \cite{aboah2022plc} upholds the motivation of this work: effective unsupervised ensembling of anomaly detection algorithms could result in a robust detection model capable of identifying anomalies in arbitrary datasets.

Stacking-based ensemble methods for classification purposes using supervised machine learning meta-detectors were examined in  \cite{xia2018novel, el2018using, akhtar2019diagnosis, javan2019intelligent, ahmed2020exploring, williams2019variable}. However, a direct comparison between the proposed stacking-based ensemble method in this work and the works in \cite{xia2018novel, el2018using, akhtar2019diagnosis, javan2019intelligent, ahmed2020exploring, williams2019variable} is unrealistic because of the unsupervised ensemble approach adopted in this work for anomaly detection purposes.

In order to validate the results for the best performing algorithm, stacking-based ensemble method is compared with previous work trained and evaluated on the same TLIGHT system dataset by using statistical hypothesis test. Statistical hypothesis testing is employed to validate and ensure fair comparison between the stacking ensemble method and previous work. Table~\ref{tab3} compares the average performance of the best algorithm in \cite{aboah2022plc} with the proposed stacking algorithm in this work. Statistical evidence about the best-performing technique is conducted using one-way analysis of variance (ANOVA) in order to determine if there is any significant difference between the average performance of the method in this work and that of previous work \cite{aboah2022plc}. F1-score is selected as an evaluation metric in the hypothesis test because it measures the balance between precision and recall. The following assumptions are made about the ICS dataset 

\begin{itemize}
\item	data points in each test sample identically distributed and independent; and
\item	data points in each test sample are normally distributed.
\end{itemize}
Additionally, the hypotheses for the statistical test are
\begin{itemize}
\item	null hypothesis $(H_0)$: The mean F1-scores of all anomaly detection models are equal; and
\item	alternate hypothesis $(H_a)$: One or more of the mean F1-score of the anomaly detection models are unequal.
\end{itemize}

Anomaly detection performance of the stacking-based ensemble in this work and the IF in \cite{aboah2022plc} are evaluated on twenty different test samples of the exact sizes as test sets 1-5. Furthermore, each model's F1-score is computed on the twenty different samples of each test set. Results from one-way ANOVA test shows an $F\ value$ of 42.54, and a p-value $<$ 0.001. Since the p-value is below 0.05, one-way ANOVA provides significant evidence to reject the null hypothesis. Rejecting the null hypothesis means there is a significant difference between IF \cite{aboah2022plc} and stacking-based ensemble with $95\%$ confidence level. Because performance comparison is between two algorithms, one-way ANOVA results are sufficient to conclude that the stacking-based ensemble method outperforms IF \cite{aboah2022plc}.

\subsection{Practical Considerations and Limitations}
The unsupervised ensemble methods presented in this work can be implemented in real-world PLC-based ICS infrastructure. One way is to adapt the experimental approach in this work, or the approach in \cite{aboah2022plc} by serializing the trained ensemble models onto a separate computer with a data pipeline to the HMI historian and PLC memory addresses to receive data and perform real-time anomaly detection. Alternatively, the proposed methods can be implemented on both legacy and embedded PLCs by adapting a similar process outlined in \cite{aboah2022plc}. That is, the trained ensemble models can be compiled to C code using open-source compilers \cite{unlu2020efficient, OliverUrbann}. The resulting C code should be readily portable to dedicated embedded or general-purpose processors \cite{urbann2020ac} for real-time anomaly detection.

Although this work proposes robust unsupervised ensemble techniques for anomaly detection in ICS, some limitations are associated with the proposed methods. First, ensembling OCSVM, OCNN, and IF base detectors make the proposed methods in this work expensive in terms of time and storage. That large amount of memory is required to store outputs of the base detectors and their ensemble output during training and testing. Moreover, the high time complexities of the proposed methods result from the effective summation of the base models' time complexities and the time required for the voting function in each method to compute the final anomaly score. Despite the stacking-based method's superior performance in all test cases, its IF meta-detector still requires hyperparameter tuning to achieve such robust performance. Lastly, the proposed WV-OLS has a fundamental limitation of high bias in situations where two or more base detectors have similar objective functions. The high bias effect is reduced by using a KNN learning algorithm for assigning the weights. In practice, a learning algorithm for the weighted voting method should be selected based on its ability to handle non-linear features and has low bias.

\section{Conclusion}
This work proposes five unsupervised ensemble anomaly detection techniques: majority-vote, maximum-score, soft voting, weighted voting, and stacking-based ensemble methods, specifically for anomaly detection in ICS. The proposed techniques incorporate OCSVM, OCNN, and IF base detectors for learning representations from the input dataset and voting strategy algorithms for combining the representations into final prediction scores. Dataset from a previously studied TLIGHT ICS system is used to train and evaluate the proposed algorithms. Majority-vote and maximum-score ensembles outperform soft-voting and WV-OLS with $5\%$ higher average accuracy, precision, and recall on all test datasets. The WV-OLS ensemble method's lower performance relative to majority vote and maximum score methods is because of OCSVM and OCNN's similar objective functions. As a result, the WV-OLS method's assigned weights are biased towards OCSVM and OCNN, leading to a biased outcome. Changing the weighted-voting learning algorithm to KNN results in an improved performance which was at par with majority-vote ensemble method. Stacking-based ensemble outperform all other ensemble methods with $10\%$ higher average accuracy, precision, recall, and F1-score on all test datasets. Unlike other ensemble methods, which struggle to detect timing bits anomalies, stacking-based ensemble has excellent detection performance on timing bits anomalies. Not only does stacking-based ensemble produce high anomaly detection performance, but the approach is confident about all predictions. A hypothesis test involving analysis of variance is used to compare stacking-based ensemble method with previous work. The hypothesis test indicates that stacking-based ensemble outperforms previous work trained on the same dataset. The high performance of the stacking-based ensemble method on all evaluation metrics compared to previous work justifies the motivation of this work: effective unsupervised ensembling of anomaly detection algorithms could result in a robust detection model capable of identifying anomalies in arbitrary ICS datasets. 

Future work would be as follows:
\begin{itemize}
\item	optimizing the weighted voting ensemble learning algorithm to mitigate its fundamental limitation of high bias in situations where two or more base detectors have similar objective functions and have correlated errors; and
\item	extending the proposed ensemble methods to detecting anomalies in ICS network layer.
\end{itemize}

\bibliographystyle{IEEEtran}
\bibliography{bare_jrnl_new_sample4}

\vspace{11pt}
\begin{IEEEbiography}[{\includegraphics[width=1in,height=1.25in,clip,keepaspectratio]{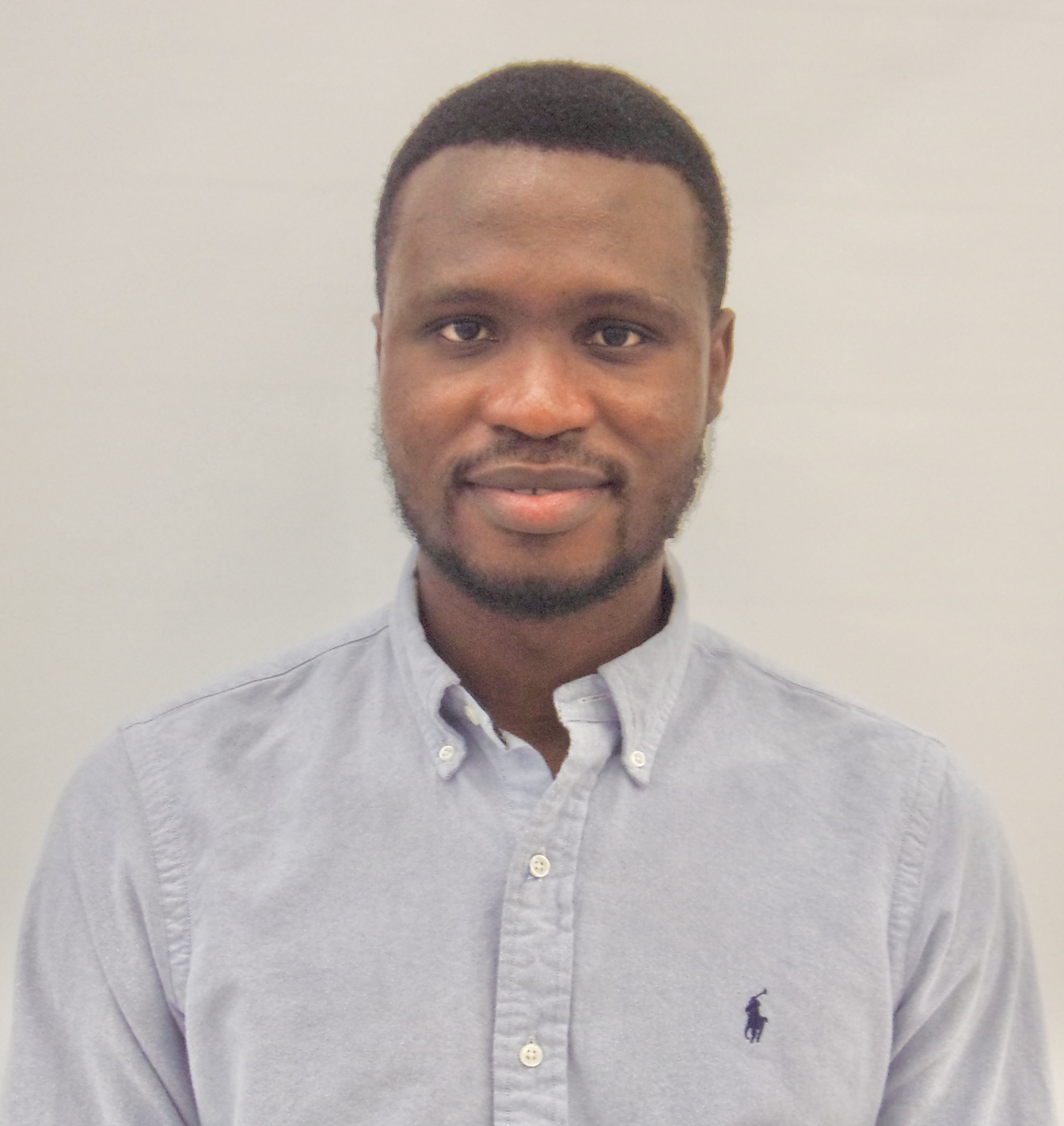}}]{\textbf{Emmanuel Aboah Boateng}} (Member, IEEE) received the BSc. degree in Electrical and Electronic Engineering from the University of Mines and Technology, Tarkwa, Ghana, in 2017 and the M.S degree in Electronics Engineering from Norfolk State University, Norfolk, VA, USA, in 2019. He is currently pursuing the Ph.D. degree in Electrical and Computer Engineering at Tennessee Technological University, USA, as a Carnegie Classification Scholar, where he specializes in developing artificial intelligence and data provenance analysis techniques to detect anomalies in hardware/embedded devices and cyber-physical systems. His research interests include machine learning, anomaly detection, cyber-physical systems security, signal processing, and fault diagnosis in hardware/embedded devices.
\end{IEEEbiography}

\begin{IEEEbiography}[{\includegraphics[width=1in,height=1.25in,clip,keepaspectratio]{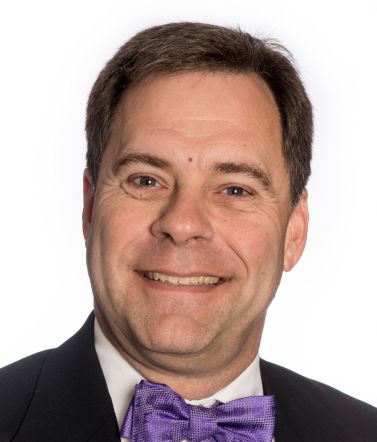}}]{J.W. Bruce} (Senior Member, 2003; Member, 1994, Student Member, 1988)
Dr. Bruce received the B.S.E. from the University of Alabama in Huntsville in 1991, the M.S.E.E. from the Georgia Institute of Technology in 1993, and the Ph.D. from the University of Nevada Las Vegas in 2000, all in electrical engineering. Dr. Bruce was a graduate research fellow of the Audio Engineering Society from 1998-2000.
In 2018, Dr. Bruce joined the faculty in the Department of Electrical \& Computer Engineering at Tennessee Technological University in Cookeville, TN.  From 2000-2018, Dr. Bruce served in the Department of Electrical and Computer Engineering at Mississippi State University.   Dr. Bruce has contributed to the research areas of embedded systems design, engineering education, UAVs, and data converter architecture design. Dr. Bruce’s research has resulted in more than 50 technical publications, one book chapter, and two books.
\end{IEEEbiography}

\end{document}